\documentclass[10pt,journal]{IEEEtran}
\usepackage{amsmath,amsfonts}
\usepackage{algorithmic}
\usepackage{algorithm}
\usepackage{caption}
\usepackage{subcaption}
\usepackage{array}
\usepackage[caption=false,font=normalsize,labelfont=sf,textfont=sf]{subfig}
\usepackage{textcomp}
\usepackage{stfloats}
\usepackage{url}
\usepackage{verbatim}
\usepackage{graphicx}
\usepackage{cite}
\usepackage{color}
\usepackage{multirow}
\usepackage{bbm}
\usepackage{colortbl}
\usepackage{tcolorbox}
\usepackage{bbding}
\usepackage{titlesec}
\usepackage{wrapfig}
\usepackage{threeparttable}
\usepackage{booktabs}
\usepackage{makecell}
\usepackage[switch]{lineno}

\begin{document}
\title{{AdaptGCD: Multi-Expert Adapter Tuning  for \\ Generalized Category Discovery}}

\author{Yuxun~Qu,  
        Yongqiang~Tang,
        Chenyang~Zhang
        and~Wensheng~Zhang
\thanks{ This work has been submitted to the IEEE for possible publication. Copyright may be transferred without notice, after which this version may no longer be accessible.}
\thanks{ Yuxun Qu is with the College of Computer Science, Nankai University, Tianjin 300350, China. (e-mail: quyuxun2022@mail.nankai.edu.cn)}
\thanks{ Yongqiang Tang and Chenyang Zhang are with the State Key Laboratory of Multimodal Artificial Intelligence Systems, Institute of Automation, Chinese Academy of Sciences, Beijing 101408, China. (e-mail: yongqiang.tang@ia.ac.cn, chenyang.zhang@ia.ac.cn)}
\thanks{ Wensheng Zhang is with the College of Computer Science, Nankai University, Tianjin 300350, China, and also with the State Key Laboratory of Multimodal Artificial Intelligence Systems, Institute of Automation, Chinese Academy of Sciences, Beijing 101408, China. (e-mail: zhangwenshengia@hotmail.com).}}

\markboth{Submit to IEEE Transactions on Circuits and Systems for Video Technology (TCSVT), 2025}%
{Shell \MakeLowercase{\textit{et al.}}: A Sample Article Using IEEEtran.cls for IEEE Journals}


\maketitle

\begin{abstract}
Different from the traditional semi-supervised learning paradigm that is constrained by the close-world assumption, Generalized Category Discovery (GCD) presumes that the unlabeled dataset contains new categories not appearing in the labeled set, and aims to not only classify old categories but also discover new categories in the unlabeled data. Existing studies on GCD typically devote to transferring the general knowledge from the self-supervised pretrained model to the target GCD task via some fine-tuning strategies, such as partial tuning and prompt learning. Nevertheless, these fine-tuning methods fail to make a sound balance between the generalization capacity of pretrained backbone and the adaptability to the GCD task. To fill this gap, in this paper, we propose a novel adapter-tuning-based method named AdaptGCD, which is the first work to introduce the adapter tuning into the GCD task and provides some key insights expected to enlighten future research. Furthermore, considering the discrepancy of supervision information between the old and new classes,  a multi-expert adapter structure equipped with a route assignment constraint is elaborately devised, such that the data from old and new classes are separated into different expert groups. Extensive experiments are conducted on 7 widely-used datasets. The remarkable performance improvements highlight the efficacy of our proposal and it can be also combined with other advanced methods like SPTNet for further enhancement.
\end{abstract}


\section{Introduction}
\label{sec:intro}

\begin{figure}[tp]
\centering
\includegraphics[width=0.45\textwidth]{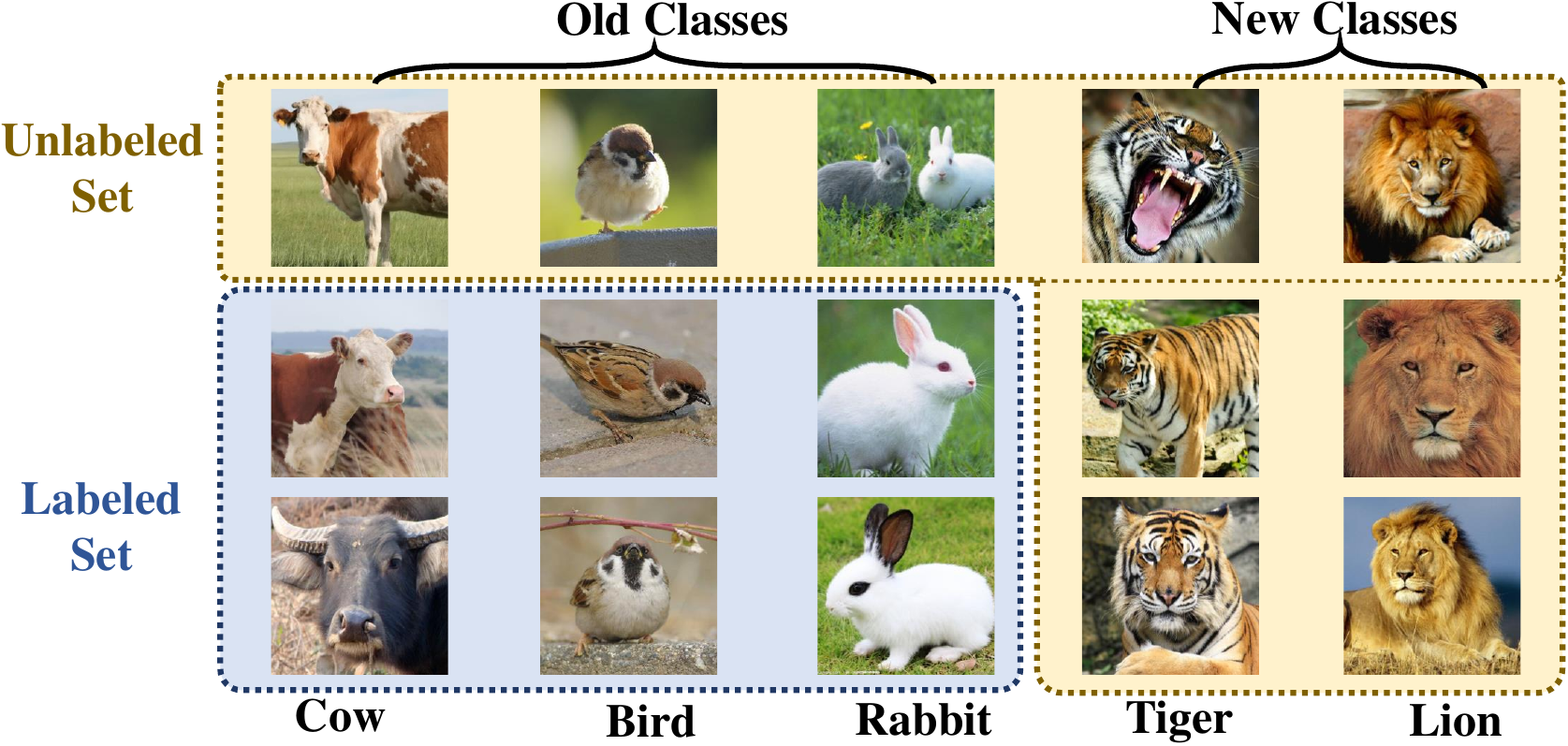}
\vspace{-3pt}
\caption{The description of the generalized category discovery (GCD) task. In this context, the unlabeled dataset contains new classes that are not present in the labeled set. }
\vspace{-0.7cm}
\label{fig::gcd} 
\end{figure}

{\color{black}Recent advancements in deep learning have been largely driven by the availability of high-quality labeled data, which incurs significant costs associated with manual annotation. To tackle this challenge, Semi-Supervised Learning (SSL) \cite{DBLP:journals/tcsv/TangCYLY24,DBLP:journals/tcsv/LuJLLLY24,DBLP:journals/tcsv/HuH24,DBLP:journals/tcsv/LiuYGY25,DBLP:journals/tcsv/JiaLWW25} has gained considerable attention as it effectively combines a limited number of labeled samples with extensive unlabeled data for model training.} While recent semi-supervised studies \cite{DBLP:conf/nips/SohnBCZZRCKL20,DBLP:conf/nips/BerthelotCGPOR19,DBLP:conf/nips/ZhangWHWWOS21,DBLP:conf/nips/LiWLYY0L23, DBLP:conf/nips/SunSL23, DBLP:conf/nips/Huang0Y0L23} show a substantial progression, they are typically constrained by the closed-world assumption, i.e., labels are provided for each category that the model needs to classify \cite{DBLP:conf/iclr/CaoBL22}. The constraint hinders the scalability of these methods. To relax this assumption, Generalized Category Discovery (GCD) \cite{DBLP:conf/cvpr/gcdVazeHVZ22} presumes that unlabeled dataset contains new categories not appearing in the labeled set, as shown in Fig. \ref{fig::gcd}. The goal of GCD is not only to classify the old categories but also to discover the new categories in the unlabeled data. {\color{black}As a promising research topic with the potential to broaden the application scope of SSL, GCD has garnered significant attention from the community in recent years.}

The primary challenge in GCD is learning the expressive representation applicable to both old and new classes \cite{DBLP:conf/cvpr/gcdVazeHVZ22}. Due to the scarcity of labels in GCD tasks, the model trained from scratch is prone to overfit the labeled data in old classes, consequently disrupting the generality of representation  \cite{wang2024sptnet}. As an alternative, the existing studies in GCD broadly fine-tune the self-supervised pretrained models, like representation learning model DINO \cite{DBLP:conf/iccv/CaronTMJMBJ21}, to transfer the strong general knowledge from the pretrained backbone to the target dataset in GCD task. In this process, a concern that is key yet often overlooked is to strike a balance between the generalization capacity from the upstream knowledge and the adaptability to the downstream GCD task, which is largely influenced by the fine-tuning strategy \cite{DBLP:journals/corr/abs-2403-14608, DBLP:conf/nips/LiuTMMHBR22,DBLP:conf/eccv/vptJiaTCCBHL22,DBLP:conf/acl/LiL20}.

Up to now, two fine-tuning strategies are widely utilized in GCD: the {partial tuning} and the prompt tuning. The straightforward {partial tuning} strategy only updates the parameters from the last few blocks while the rest parameters in the backbone are frozen \cite{DBLP:conf/cvpr/gcdVazeHVZ22}. Such approaches potentially impair the generalization ability with limited annotations since the pretrained knowledge embodied in the backbone is disrupted. Conversely, the prompt tuning methods \cite{DBLP:conf/eccv/vptJiaTCCBHL22,DBLP:conf/nips/WangC0H23} like PromptCAL \cite{DBLP:conf/cvpr/promptcalgcdZhangKSNCK23} and SPTNet \cite{wang2024sptnet} freeze all network parameters in the backbone and incorporate a learnable prompt at the input for training. Although effectively preserving the knowledge of the pretrained network, unfortunately, it is demonstrated that the prompt-based approaches are not robust to the bad data quality \cite{wang2024sptnet}, and at the same time present limited adaptability when the downstream task differs from the upstream pretraining task \cite{DBLP:conf/eccv/vptJiaTCCBHL22}.
This is particularly evident during the adaptation process from representation learning methods to the downstream classification task, resulting in the performance on GCD tasks far from satisfactory.


Luckily, as one of the most characteristic fine-tuning approaches, adapter tuning \cite{DBLP:conf/nips/adaptformerChenGTWSWL22, DBLP:conf/nips/RebuffiBV17,DBLP:conf/nips/LeiBBALZ0ZWLZC23,DBLP:conf/nips/Page-CacciaPSPR23,DBLP:conf/nips/ChengSXW0GS23} may provide a promising solution to the issues mentioned above. The adapter tuning leaves the {\color{black}parameters in} pretrained model fixed and brings in a modest number of task-specific parameters via bottleneck adapter modules \cite{DBLP:journals/corr/gloraabs-2306-07967}. Obviously, on mechanism, the adapter naturally fits for the GCD task. It could preserve pretrained knowledge nondestructively by maintaining the parameters in the backbone \cite{DBLP:conf/acl/WangTDWHJCJZ21, DBLP:journals/taslp/HouZWWQXS22, DBLP:conf/miccai/ZhangHZZZW23}. Additionally, previous work has demonstrated its capability to adapt from upstream feature learning tasks to various downstream tasks \cite{DBLP:conf/nips/adaptformerChenGTWSWL22, DBLP:journals/corr/abs-2302-08106, DBLP:conf/acl/Ye020, DBLP:conf/cvpr/LiLB22, DBLP:conf/aaai/XinDWLY24}. Surprisingly, the adapter has not yet attracted much attention on GCD task. When introducing adapter into GCD, an additional concern arises accordingly, i.e., the information imbalance between old and new classes within adapter \cite{DBLP:conf/iclr/CaoBL22}. Due to the existence of supervised loss for labeled data, the adapter receives more supervision from old classes than the new ones, thereby resulting in a bias towards old classes. Under this framework, how to render old classes and new classes non-interfering in GCD task deserves further exploration.

In this study, we propose a novel multi-expert adapter tuning method AdaptGCD for generalized category discovery. Unlike the previous partial tuning or prompt-like methods, AdaptGCD integrates learnable adapters parallel to the feed-forward layer of the ViT block and keeps the parameters from the backbone frozen to preserve the pretrained knowledge. Considering the discrepancy of supervision information between the old and new classes, the adapter is further extended to a Multi-Expert Adapter structure (MEA), which encompasses multiple adapter experts and employs a routing layer to schedule the usage of these experts. In order to guide the experts to specialize in data from the old or new classes, a route assignment constraint is further designed.
It is also noted that the proposal could act as a plugin, allowing integration with other cutting-edge methods like SPTNet, to achieve the superior performance. In summary, the key contributions are listed as follows: 
\begin{itemize}
    \item Our proposal is the first work to introduce the adapter tuning into the GCD task. Such combination is elaborate, as it mechanistically allows preserving the general knowledge of pretrained backbone and meanwhile enhancing the adaptability to the downstream GCD task. {It can also serve as a plug-and-play module to further improve the performance of the advanced method.} We believe this could provide crucial insight for the future work of GCD.
    \item To reduce the information interference between old and new classes, we employ a multi-expert adapter structure and design a route assignment constraint to separate the data from old and new classes into different adapter expert groups, {\color{black}alleviating the bias towards old classes}. 
    \item We conduct extensive experiments on 7 popular datasets. Impressively, the adapter improves the GCD performance by a large margin, which highlights the applicability of adapter tuning on the GCD task. Furthermore, the results also confirm the effectiveness of multi-expert adapter structure with route assignment constraint.
\end{itemize}

The rest of the paper is organized as follows. First, in Section \ref{section:related work}, we provide a brief review of related work. Next, Section \ref{sec:preliminary} introduces the widely used preliminary in the GCD task. In Section \ref{section:proposal}, we elaborate on the details of the proposed AdaptGCD framework. Subsequently, extensive experiments are conducted, and additional experimental analyses are presented in Section \ref{section:experiment}. Finally, we summarize the paper in Section \ref{section:conclusion}.

\section{Related Work} \label{section:related work}

\subsection{Semi-Supervised Learning}
To date, advancements in deep learning have largely depended on high-quality labeled data, which incurs substantial costs due to manual annotation. To reduce this reliance, researchers have turned to semi-supervised learning (SSL) \cite{DBLP:journals/tcsv/ChenWWPIL18, DBLP:journals/tcsv/QiuNXQX19, DBLP:journals/tcsv/0091K0H21,DBLP:journals/tcsv/ZhuG22,DBLP:journals/tcsv/XuXSZLLZ22,DBLP:journals/tcsv/WangFZWL22,DBLP:journals/tcsv/LiuSDZWZM22,DBLP:journals/tcsv/LiLS22,DBLP:journals/tcsv/JiaLLH23,DBLP:journals/tcsv/TangCYLY24,DBLP:journals/tcsv/LuJLLLY24,DBLP:journals/tcsv/HuH24,DBLP:journals/tcsv/LiuYGY25,DBLP:journals/tcsv/JiaLWW25}, which leverages a limited number of labeled samples alongside extensive unlabeled datasets for model training. Modern SSL algorithms primarily consist of two paradigms: consistency regularization \cite{DBLP:conf/iclr/LaineA17, DBLP:conf/nips/TarvainenV17} and pseudo-labeling methods \cite{DBLP:conf/ijcnn/ArazoOAOM20, DBLP:conf/iclr/RizveDRS21}. Consistency regularization requires the model to learn consistent representations or predictions of the same image across various augmentations or training phases, with notable works including temporal ensembling \cite{DBLP:conf/iclr/LaineA17} and mean teacher \cite{DBLP:conf/nips/TarvainenV17}. In contrast, pseudo-labeling methods \cite{DBLP:conf/ijcnn/ArazoOAOM20} estimate reliable pseudo-labels for unlabeled data. Recently, FixMatch \cite{DBLP:conf/nips/SohnBCZZRCKL20} has emerged as a combination of these paradigms, estimating pseudo-labels from weakly augmented images and minimizing cross-entropy loss between these pseudo-labels and predictions from strongly augmented images. This approach has garnered attention, leading to subsequent works like MixMatch \cite{DBLP:conf/nips/BerthelotCGPOR19} and FlexMatch \cite{DBLP:conf/nips/ZhangWHWWOS21}, which enhance the scalability and performance. Additionally, another promising direction is to combine contrastive learning \cite{DBLP:conf/icml/ChenK0H20} with semi-supervised learning to enhance model generalization, thereby making better use of unlabeled data. Despite progress in SSL, the mainstream SSL methods still maintain a closed-world assumption, presuming that labeled instances are available for all possible categories in the unlabeled dataset, which limits their application in more realistic scenarios.

\subsection{Generalized Category Discovery} 
As an extension of semi-supervised learning in open scenarios, Generalized Category Discovery (GCD, \cite{DBLP:conf/cvpr/gcdVazeHVZ22,DBLP:conf/nips/VazeVZ23, DBLP:conf/nips/RastegarDS23, DBLP:conf/cvpr/promptcalgcdZhangKSNCK23,wang2024sptnet, DBLP:conf/nips/BaiLWCMLZFWH23,DBLP:conf/cvpr/CaoZWYS0L024}), also known as open-world semi-supervised learning, does not assume that the unlabeled dataset comes from the same class set as the labeled dataset, posing a greater challenge for designing an effective model to classify unlabeled images across both old and new classes. The pioneering work \cite{DBLP:conf/cvpr/gcdVazeHVZ22} achieves this goal by refining representations from the pretrained model using contrastive learning and assigning classification results through semi-supervised $k$-means clustering. Some concurrent works \cite{DBLP:conf/cvpr/PuZS23, DBLP:conf/iccv/gmmgcdZhaoW023, DBLP:conf/cvpr/promptcalgcdZhangKSNCK23} also follow the contrastive representation learning paradigm. Distinct from these methods, the parametric approach SimGCD \cite{DBLP:conf/iccv/simgcdWenZQ23} introduces a learnable classifier and self-distillation to supervise the classifier, achieving boosting performance. Furthermore, subsequent works have made further improvements to solve the issues of forgetting during training \cite{DBLP:conf/cvpr/CaoZWYS0L024} and the desynchronization of teacher-student network knowledge \cite{DBLP:conf/nips/LinAWCTWWD024}. Despite excellent performance, most of these studies \cite{DBLP:conf/cvpr/gcdVazeHVZ22,DBLP:conf/iccv/simgcdWenZQ23} adopt simple fine-tuning strategies that only update the last block of the vision transformer, which disrupts pretrained knowledge. Recently emerged prompt-based methods, such as PromptCAL \cite{DBLP:conf/cvpr/promptcalgcdZhangKSNCK23} and SPTNet \cite{wang2024sptnet}, can alleviate the disruption of backbone to some extent, but they still lack adaptability to various tasks and datasets. Hence, this work is devoted to seeking an effective fine-tuning strategy for the GCD task to preserve priority and enhance adaptability to downstream tasks simultaneously.

\subsection{Fine-Tuning Strategy}
The mainstream fine-tuning approaches include partial tuning, prompt tuning and adapter tuning methods. Partial tuning methods freeze most of the backbone and fine-tune a small portion of parameters, like linear heads \cite{DBLP:conf/cvpr/IofinovaPKA22}, MLP heads \cite{DBLP:journals/corr/abs-2003-04297}, or several blocks of the backbone \cite{DBLP:conf/cvpr/HeCXLDG22}. Although this strategy is widely used in GCD, it still falls short of maintaining the network's generalization capabilities. Prompt tuning \cite{DBLP:conf/eccv/vptJiaTCCBHL22, DBLP:conf/iccv/E2VPTHanWCCWQL23} prepends a set of learnable vectors to the input and only updates these prompts during fine-tuning. Despite their admirable performance, these methods present limited adaptability when a large gap exists between the downstream task and the upstream pretraining task \cite{DBLP:conf/eccv/vptJiaTCCBHL22}, especially during the adaptation from upstream representation learning tasks to downstream classification tasks. Adapter tuning, including AdaptFormer \cite{DBLP:conf/nips/adaptformerChenGTWSWL22}, RepAdapter \cite{DBLP:journals/corr/abs-2302-08106}, etc., integrates shallow modules with the fixed backbone to protect the priority in pretrained models. Recent works \cite{DBLP:conf/acl/Ye020,DBLP:conf/cvpr/LiLB22, DBLP:conf/aaai/XinDWLY24} also show its flexibility to adapt to different downstream tasks. Considering the excellent mechanism of adapter tuning, this work further concentrates on the application and modification of the adapter-tuning-based strategies in GCD.

\subsection{Mixture of Expert}
The Mixture of Experts (MoE) is a machine learning technique employing route layer to segment a single task space into multiple subtasks \cite{DBLP:journals/corr/abs-2402-08562,DBLP:journals/corr/abs-2311-03285,DBLP:journals/corr/abs-2310-18339,DBLP:journals/corr/abs-2310-18547,DBLP:journals/corr/abs-2307-13269}. Each subtask is specifically managed by individual expert networks and the prediction result is aggregation of the solutions provided by these experts. In recent years, there appear some studies focusing on the fusion of the MoE with Adapters or LoRA \cite{DBLP:journals/corr/abs-2310-18339,dou2024loramoe}. For instance, LoRAMoE \cite{dou2024loramoe} introduces a MoE-style plugin and a localized balancing constraint to segregate world knowledge and downstream information into two distinct sets of experts.
Inspired by LoRAMoE, the proposed approach also introduces multi-expert technology to handle data from new and old classes in diverse experts, thus reducing interference during the learning process.

\section{Preliminary}
\label{sec:preliminary}
In this section, we will introduce the basic setting for the generalized category discovery task and then elaborate on the pipeline of the baseline SimGCD.
\subsection{Setting of Generalized Category Discovery}

Assume that the training dataset is composed of the labeled and unlabeled parts defined as $\mathcal{D}^l = \{(\mathbf{x}^l_i, y^l_i)\} \in \mathcal{X} \times \mathcal{Y}^l$ and $\mathcal{D}^u = \{(\textbf{x}^u_i, y^u_i )\} \in \mathcal{X}\times \mathcal{Y}^u$, where $\mathcal{Y}^l, \mathcal{Y}^u$ are the label spaces of the labeled and unlabeled samples. {$K=|\mathcal{Y}^u|$ is the total number of categories.} In GCD setting, $\mathcal{Y}^l \subset \mathcal{Y}^u$. Given $\mathcal{Y}^n=\mathcal{Y}^u \backslash \mathcal{Y}^l$, the categories in the $\mathcal{Y}^l$ are denoted as the ``old'' classes while those in $\mathcal{Y}^n$ are the ``new'' classes. The ultimate goal of the generalized category discovery task is to learn a model to categorize the samples from $\mathcal{D}^u$ in both the old and new classes.

\subsection{Baseline: SimGCD}

As a simple yet effective parametric approach in the field of GCD, SimGCD \cite{DBLP:conf/iccv/simgcdWenZQ23} is adopted as the baseline of the proposed method. It incorporates two vital losses, including the representation learning loss $\mathcal{L}_{\rm rep}$ and the classification objective $\mathcal{L}_{\rm cls}$.

\begin{figure*}[!htbp]
\center
\includegraphics[width=0.8\textwidth]{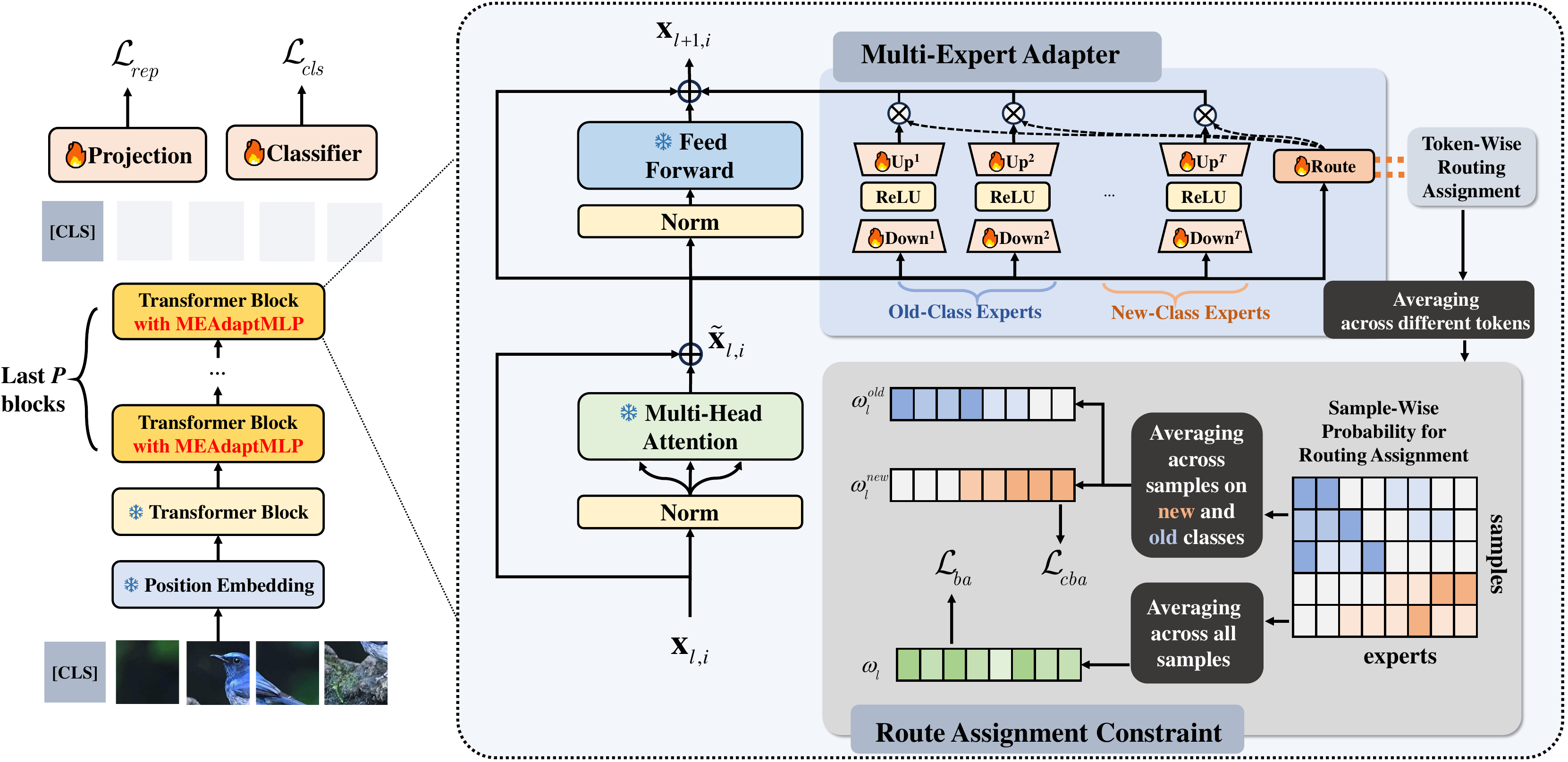}
\caption{{The framework of our proposed AdaptGCD.} It contains two critical modules: the multi-expert adapter (MEA) structure and the route assignment constraint. The MEA introduces multiple adapter experts and the route assignment constraint, {\color{black}including the balanced assignment loss $\mathcal{L}_{ba}$ and category-aware balanced assignment loss $\mathcal{L}_{cba}$,} guides the allocation of these experts. } 
\vspace{-0.5cm}
\label{fig::framework} 
\end{figure*}

\emph{1) Representation Learning Loss:} The representation learning loss integrates supervised contrastive learning \cite{DBLP:journals/corr/abs-2004-11362} for labeled samples with self-supervised contrastive learning \cite{DBLP:conf/icml/ChenK0H20} for all samples. Given two random augmented views of the same image $\textbf{x}_i$ and $\textbf{x}'_i$ in a mini-batch $\mathcal{B}$, the self-supervised contrastive loss $\mathcal{L}_{\rm rep}^u$ and supervised contrastive loss $\mathcal{L}_{\rm rep}^l$ are represented as Eq. (\ref{cal::reploss}).
\begin{equation}
    \label{cal::reploss}
    \begin{aligned}
    \mathcal{L}_{\rm rep}^u =& \frac{1}{|\mathcal{B}|}\sum_{i\in \mathcal{B}} -\log \frac{e^{\textbf{z}^{\top}_{i} \textbf{z}'_{i}/\tau_u}}{\sum_{i}^{i\neq n}{e^{\textbf{z}^{\top}_i \textbf{z}'_n / \tau_u}}},\\
    \mathcal{L}_{\rm rep}^s =& \frac{1}{|\mathcal{B}^{l}|}\sum_{i\in \mathcal{B}^{l}} \sum_{q\in |\mathcal{N}_{i}|} -\log \frac{e^{\textbf{z}^{\top}_{i} \textbf{z}'_{q}/\tau_{c}}}{\sum_{i}^{i\neq n}{e^{\textbf{z}^{\top}_{i} \textbf{z}'_{n} / \tau_{c}}}},
    \end{aligned}    
\end{equation}
where $\textbf{z}_i=g(f(\textbf{x}_i))$ represents $\ell_2$-normalised feature, $f(\cdot)$, $g(\cdot)$ denote the backbone and projection head, respectively. The hyper-parameter $\tau_u,\tau_c$ symbolize temperature values, $\mathcal{N}_i$ indicates all other labeled data with the same label as $\textbf{x}_i$ in the same batch, and $\mathcal{B}_l$ corresponds to the labeled subset of $\mathcal{B}$. Subsequently, the total loss in representation learning merges the two components with a balancing factor $\lambda$, denoted by $\mathcal{L}_{rep} = (1 - \lambda) \mathcal{L}_{rep}^u + \lambda \mathcal{L}_{rep}^s$.

\emph{2) Classification Objective:} SimGCD employs the parametric classifier in conjunction with the self-distillation paradigm for the learning of the classifier. Given the set of category prototypes $\mathcal{C}=\{\textbf{c}_1,...\textbf{c}_K\}$, where each one corresponds to a class and $K=|\mathcal{Y}^u|$ is the total number of categories. As depicted in Eq. (\ref{equ::prediction}), the prediction is determined by the cosine distance between the hidden features $\textbf{h}_i=f(\textbf{x}_i)$ and the class prototypes.
\begin{equation}
    \begin{aligned}
    \label{equ::prediction}
    \textbf{p}_i^{(k)} = \frac{\exp\left( \frac{1}{\tau_s} (\textbf{h}_i/||\textbf{h}_i||_2)^{\top}(\textbf{c}_k/||\textbf{c}_k||_2)\right)}{\sum_{k'}\exp\left( \frac{1}{\tau_s} (\textbf{h}_i/||\textbf{h}_i||_2)^{\top}(\textbf{c}_{k'}/||\textbf{c}_{k'}||_2)\right)}.
    \end{aligned}    
\end{equation}
In a similar manner, the soft pseudo-label $\textbf{q}'_i$ is generated using another view $\textbf{x}'_i$ with a sharper temperature $\tau_r$. Subsequently, the classification losses are defined as the cross-entropy loss between the predictions and pseudo-labels or ground-truth labels. The formulation is represented in Eq. (\ref{cal::clsloss}).
\begin{equation}
    \label{cal::clsloss}
    \begin{aligned}
    \mathcal{L}_{\rm cls}^u =& \frac{1}{|\mathcal{B}|} \sum_{i\in |\mathcal{B}|} \ell(\textbf{q}'_i,\textbf{p}_i) - \epsilon H(\overline{\textbf{p}}),\\
    \mathcal{L}_{cls}^s = &\frac{1}{|\mathcal{B}^l|} \sum_{i\in \mathcal{B}^l} \ell(\sigma({y}_i),\textbf{p}_i),
    \end{aligned}    
\end{equation}
where $\ell(\cdot,\cdot)$ refers to the cross entropy loss, $\sigma(\cdot)$ denote the one-hot operation for labels, entropy regularizer $H(\bar{\textbf{p}}) = - \sum_{k} \bar{\textbf{p}}^{(k)}\log \bar{\textbf{p}}^{(k)}$ and $\bar{p} = \frac{1}{2|\mathcal{B}|}\sum_{i\in \mathcal{B}}(\textbf{p}_i+\textbf{p}'_i)$. These elements are integrated into the classification objective as $\mathcal{L}_{cls} = (1 - \lambda) \mathcal{L}_{\rm cls}^u + \lambda \mathcal{L}_{\rm cls}^s$. All in all, the objective for SimGCD is described as $\mathcal{L}_{sgcd}=\mathcal{L}_{\rm rep} + \mathcal{L}_{\rm cls}$.

\emph{3) Partial Tuning:} The effectiveness of the SimGCD framework largely stems from pretrained model. To take advantage of the generic representation from upstream tasks (like DINO \cite{DBLP:conf/iccv/CaronTMJMBJ21}), the SimGCD framework only unfreezes the pretrained parameters of the last block for fine-tuning. However, as mentioned previously, this method might still result in information loss from the pretrained backbone, thereby reducing the generality of features.

\section{The Proposed Method} \label{section:proposal}

In this work, a novel AdaptGCD framework (as depicted in Fig. \ref{fig::framework}) is proposed for tackling the generalized category discovery task. It contains two critical modules, including the multi-expert adapter structure and the route assignment constraint. In the following contents, we would elaborate on our proposed method.

\subsection{Multi-Expert Adapter} 
In this part, the proposed approach integrates the adapter structure into the GCD framework, introducing a learnable additional architecture into the backbone while keeping the rest parameters in the backbone frozen during training.  Then the basic adapter is further developed into the multi-expert adapter, allowing data under different supervisions to navigate through diverse paths.

\emph{1) Basic Adapter:} The designation of the adapter in this work follows the basic architecture proposed by AdaptFormer \cite{DBLP:conf/nips/adaptformerChenGTWSWL22}, which replaces the Feed-Forward Network (FFN) in the Transformer with the AdaptMLP module. This module contains two parallel sub-branches, one is identical to the Feed-Forward block while the other is an learnable lightweight bottleneck module. The bottleneck module includes a down-projection layer with parameters 
$\textbf{W}_{\rm down} \in \mathbb{R}^{\hat{d}\times d}$, $\textbf{b}_{\rm down} \in \mathbb{R}^{1\times\hat{d}}$
and an up-projection layer $\textbf{W}_{\rm up} \in \mathbb{R}^{d\times \hat{d} }$, $\textbf{b}_{\rm up} \in \mathbb{R}^{1\times{d}}$, where $d$ and $\hat{d}$ denote the input dimension of FFN and middle dimension of bottleneck ($\hat{d}<d$). The down-projection layer and the up-projection layer are connected with the ReLU layer for nonlinear properties, and the entire bottleneck module is linked to the original FFN with the residual connection, as depicted in Fig. \ref{fig::framework}. Specifically, for the features of the $i$-th sample fed into FFN of the $l$-th Transformer blocks $\widetilde{\textbf{x}}_{l,i}$, the forward process of bottleneck module to yield the adapted feature $\Delta{\textbf{x}}_{l,i}$ is formulated as follows:
\begin{equation}
    \label{cal::bottleneck}
    \begin{aligned}
        \Delta{\textbf{x}}_{l,i} = s \left({\rm ReLU}(\widetilde{\textbf{x}}_{l,i} \textbf{W}_{\rm down}^{\top} + \textbf{b}_{\rm down}) \textbf{W}_{\rm up}^{\top} + \textbf{b}_{\rm up}\right),
    \end{aligned}
\end{equation}
where $s$ is the scale factor. Then,  the adapted features are fused with the original FFN branch by residual connection:
\begin{equation}
    \begin{aligned}
        {\textbf{x}}_{l+1,i} = \overbrace{{\rm MLP}({\rm LN}(\widetilde{\textbf{x}}_{l,i})) + \widetilde{\textbf{x}}_{l,i}}^{\text{Original FFN}}+  \overbrace{\Delta{\textbf{x}}_{l,i}}^{\text{Bottleneck}},
    \end{aligned}
\end{equation}
where $\textbf{x}_{l+1,i}$ is the output of the $l$-th Transformer block, LN($\cdot$) is the layer normalization function. 
During fine-tuning, only the parameters of the bottleneck modules are updated and the rest are frozen.

\emph{2) Multiple-Expert Adapters:} As previously mentioned \cite{DBLP:conf/iclr/CaoBL22}, data from older classes yields more supervisory information. It would result in the preference for the old class as the network receives more gradients from the old-class data than the new-class data. To counteract this implicit bias, the strategy of multi-expert technology is adopted to integrate with the adapter. This approach assigns different types of data to different expert adapters, therefore achieving separation for the data at the network level and alleviating the interference between new classes and old classes. 

Multiple-expert adapter (MEA) consists of $T$ experts for handling different features and a route function to assign weights to these experts. More specifically, for the input feature $\widetilde{\textbf{x}}_{l,i}$, MEA leverages the following route function  to estimate the weight $\boldsymbol{\omega}_{l,i}$ of multiple experts:
\begin{equation}
    \label{cal::gate}
        \begin{aligned}   
        \boldsymbol{\omega}_{l,i} = \left[\omega^1_{l,i},...\omega^t_{l,i},...,\omega^T_{l,i}\right] = & {\rm Softmax}(\widetilde{\textbf{x}}_{l,i} \textbf{W}_{\rm route}^{\top}/\tau_r),
        \end{aligned}
\end{equation}
where 
$\textbf{W}_{\rm route}\in\mathbb{R}^{T\times d}$ denotes the trainable parameters in the route function and $\tau_r$ indicates the gating temperatures. The forward process of the bottleneck module is rectified as Eq.(\ref{cal::gate_bottle}):
\begin{equation}
    \label{cal::gate_bottle}
        \begin{aligned}
            \Delta{\textbf{x}}_{l,i} = s \sum_{t=1}^{T}\omega^{t}_{l,i}\left({\rm ReLU}(\widetilde{\textbf{x}}_{l,i} \textbf{W}_{\rm down}^{t} + \textbf{b}_{\rm down}^{t})  \textbf{W}_{\rm up}^{t} + \textbf{b}_{\rm up}^{t}\right),
    \end{aligned}
\end{equation}
where $\textbf{W}_{\rm down}^{t}, \textbf{b}_{\rm down}^{t}, \textbf{W}_{\rm up}^{t}$ and $ \textbf{b}_{\rm up}^{t}$ denotes the parameters of the $t$-expert, respectively. More detailed implementation of MEA is in the Appendix. By this, the multi-expert mechanism enables the network to develop the powerful capabilities to flexibly handle the data from new and old classes. According to the previous research \cite{DBLP:journals/corr/abs-2402-08562}, allocating more adapter experts in higher blocks enhances the effectiveness of models compared with inserting them in the lower blocks. Hence, the MEA structure is incorporated only in the last $P$ blocks out of the total $L$ blocks.

\subsection{Route Assignment Constraint} 
In the multi-expert adapter, the route assignment constraint is required to supervise and control the route distribution. The assignment focuses mainly on two aspects: First, for all data, the assignment for all experts needs to be balanced to make full use of the resources of experts. Second, for data in old or new classes, the constraint assigns the corresponding experts to them so that the data can be well separated at the routing level. These two aspects correspond to the balanced assignment loss and the category-aware balanced assignment loss, which are introduced in this part. 
The proposal applies an assignment constraint on the average expert assignment of new and old class samples, rather than on individual samples as in the constraint from previous works \cite{DBLP:journals/corr/loramoeabs-2312-09979}. This enables more flexible expert assignments and reduces sensitivity to inaccurate pseudo labels. {\color{black}More comparisons between the proposed route assignment constraint and the loss in previous works are depicted in the Appendix.}

\emph{1) Sample-Wise Probability for Route Assignment:} It is worth noting that, in order to obtain the sample-wise route assignment probability, it is necessary to combine the gate vectors of different tokens for the same sample. Assume that $\boldsymbol{\omega}_{l,i,v}$ denotes the gate probability from the $v$-th token of the $i$-th sample within the $l$-th blocks, then the route assignment probability for the $i$-th sample ${\boldsymbol{\omega}}_{l,i}$ can be formulated as $ {\boldsymbol{\omega}}_{l,i} =  {\rm Softmax}\left(\frac{1}{V}\sum_{v=1}^{V} \boldsymbol{\omega}_{l,i,v}/\tau_g\right)$, where $V$ denotes the number of tokens and $\tau_g$ is the routing temperature, and ${\rm Softmax}(\cdot)$ is utilized to sharpen the gate vectors.

\emph{2) Balanced {Assignment} Loss:} The balanced {assignment} loss is designed to ensure the maximal usage of diverse experts. Previous research on mixture-of-expert \cite{DBLP:journals/corr/loramoeabs-2312-09979} demonstrated a tendency towards a ``winner-takes-all'' \cite{DBLP:journals/corr/loramoeabs-2312-09979} situation, which indicates that few experts are frequently activated and domain the learning process, resulting in the waste of some experts. To mitigate this, in this study, the mean distribution of experts in the $l$-th layer ${\boldsymbol{\omega}}_{l}$, averaging the route assignment probabilities across all samples, is aligned to a uniform distribution $I$, where $I(t)=1/T, t=1,...T$. The formulation is depicted in Eq. (\ref{cal::uniform_constraint}):
\begin{equation}
\label{cal::uniform_constraint}
    \begin{aligned}
        {\boldsymbol{\omega}}_{l} = &\frac{1}{|\mathcal{B}|} \sum_{i\in \mathcal{B}}{\boldsymbol{\omega}}_{l,i},\\
        \mathcal{L}_{ba} = & \sum_{l=L-P+1}^{L}\mathbb{D}_{\rm KL}\left({\boldsymbol{\omega}}_{l}|| {I}\right),
    \end{aligned}
\end{equation}
where $\mathcal{B}$ indicates the mini-batch and $\mathbb{D}_{\rm KL}(\cdot || \cdot)$ indicates the Kullback-Leibler divergence.

\emph{3) Category-Aware Balanced Assignment Loss:} Considering the imbalance of supervision between new classes and old classes, we propose a category-aware balanced assignment loss to separate new-class and old-class data into different experts and reduce their interference. Since the exact class label of the samples is not determined during training, the pseudo labels are utilized as a substitute. As shown in Eq. (\ref{cal::pseudo_label}), pseudo labels of the samples are obtained according to the predictions.
\begin{equation}
    \label{cal::pseudo_label}
    \begin{aligned}
        \hat{y}_i = \begin{cases}
            y_i, & \textbf{x}_i \text{~is~labeled}, \\
            {\rm argmax}_k({p}_{i,k}), & \textbf{x}_i \text{~is~unlabeled},
        \end{cases}
    \end{aligned}
\end{equation}
where $p_{i,k}$ indicates the predictive probability of $i$-sample in $k$-class. Then the route assignment probabilities of the experts for the old and new classes are estimated as ${\boldsymbol{\omega}}^{old}_{l}$ and ${\boldsymbol{\omega}}^{new}_{l}$ in Eq. (\ref{cal::conditional_prob}). 
\begin{equation}
    \label{cal::conditional_prob}
        \begin{aligned}
        {\boldsymbol{\omega}}^{old}_{l} = &\frac{\sum_i {\boldsymbol{\omega}}_{l,i}\mathbbm{1}(\hat{y}_i\in \mathcal{Y}_{l})}{\sum_i \mathbbm{1}(\hat{y}_i\in \mathcal{Y}_{l})},\\
        {\boldsymbol{\omega}}^{new}_{l} = & \frac{\sum_i {\boldsymbol{\omega}}_{l,i}\mathbbm{1}(\hat{y}_i\in \mathcal{Y}_{n})}{\sum_i \mathbbm{1}(\hat{y}_i\in \mathcal{Y}_{n})}.  
        \end{aligned}
\end{equation}
We manually specify the expert groups for the old and new classes beforehand and denote the expert groups as $\mathcal{T}^{old}$ and $\mathcal{T}^{new}$, respectively.  For instance, the first four experts are assigned to $\mathcal{T}^{old}$ and the remaining experts are naturally divided into $\mathcal{T}^{new}$.
Then the target route distribution probability for the old classes  $I^{old}\in \mathbb{R}^T$ and new classes $I^{new}\in \mathbb{R}^T$ is established as follows:
\begin{equation}
    \label{cal::target}
        \begin{aligned}
            I^{old}(t) = & \begin{cases}
                \frac{1}{|\mathcal{T}^{old}|} & t\in\mathcal{T}^{old} \\
                0 & t\in\mathcal{T}^{new} \\
            \end{cases}, \\
            I^{new}(t) = & \begin{cases}
                0 & t\in\mathcal{T}^{old} \\
                \frac{1}{|\mathcal{T}^{new}|} & t\in\mathcal{T}^{new} \\
            \end{cases}. 
        \end{aligned}
\end{equation}
These two distribution functions respectively describe the target assignment distribution of data in the old and new data. 
{\color{black}
To enhance the separation of information between new and old classes, we employ the Kullback-Leibler divergence loss as a constraint, as shown in Eq. (\ref{cal::cbl}), to align the average probability of old and new classes ${\boldsymbol{\omega}}^{old}_{l}$ and ${\boldsymbol{\omega}}^{new}_{l}$ with the predefined targeted distribution $I^{old}$ and $I^{new}$. Unlike previous works \cite{dou2024loramoe} using weighted variance coefficients as a soft constraints to balance expert allocation within samples, this strategy emphasizes isolating average expert allocation of new class samples from old class samples through distribution alignment.
}
\begin{equation}
\label{cal::cbl}
        \begin{aligned}
            \mathcal{L}_{cba} = \sum_{l=L-1+P}^{L}\left(\mathbb{D}_{\rm KL}\left({\boldsymbol{\omega}}^{old}_{l}|| {I}^{old}\right) +  \mathbb{D}_{\rm KL}\left({\boldsymbol{\omega}}^{new}_{l}|| {I}^{new}\right) \right).
        \end{aligned}
\end{equation}
In the final step, the route assignment loss $\mathcal{L}_{ra}$ for the AdaptGCD is collected as the weighted sum of the two losses, i.e. $\mathcal{L}_{ra} = \beta \mathcal{L}_{ba}+ \alpha \mathcal{L}_{cba}$, where $\alpha,\beta$ are the balancing factors. 

\emph{4) Overall Loss:} The overall loss of AdaptGCD is collected as the sum of SimGCD loss and the route assignment loss, as formulated in Eq. (\ref{cal::ovelall_loss}):
\begin{equation}
\label{cal::ovelall_loss}
    \begin{aligned}
        \mathcal{L}_{overall} = \mathcal{L}_{sgcd} + \mathcal{L}_{ra}.
    \end{aligned}
\end{equation}
The proposal could also serve as a plug-and-play component for integration with other methods. In this study, we combine it with the advanced method SPTNet \cite{wang2024sptnet}. Specifically, SPTNet consists of two training stages that learn the parameters from the prompt and the backbone, respectively. To enhance its performance, we replace the second stage, which trains the backbone using partial tuning, with the proposed AdaptGCD training strategy.

\section{Experiments} \label{section:experiment}

\subsection{Experimental Setup}\label{sec::experiment_setup}

\emph{1) Datasets:} The effectiveness of the proposed method is validated on three characteristic benchmarks: the generic image recognition benchmark including CIFAR10/100 \cite{krizhevsky2009learning} and ImageNet-100 \cite{DBLP:conf/eccv/TianKI20}, the semantic shift benchmark including CUB-200 \cite{welinder2010caltech}, Stanford Cars \cite{krause20133d}, and FGVC-Aircraft \cite{DBLP:journals/corr/MajiRKBV13} and the harder long-tailed dataset Herbarium 19 \cite{tan2019herbarium}. Following SimGCD \cite{DBLP:conf/iccv/simgcdWenZQ23}, a subset of all the classes $\mathcal{Y}^l$ is sampled as the old classes while 50\% of the images from the old classes are selected to construct the labeled set $\mathcal{D}^l$. The rest of the images are collected as the unlabeled set $\mathcal{D}^u$. Furthermore, in CIFAR100, 80\% of categories are assigned as old classes while the rest are considered as new ones. For the rest datasets, the portation of old classes is 50\%. The basic settings of the datasets \cite{DBLP:conf/cvpr/gcdVazeHVZ22} are displayed in Table \ref{table:dataset_setting}.

\emph{2) Implementation Details:
} Following GCD \cite{DBLP:conf/cvpr/gcdVazeHVZ22}, the proposed method is trained with a ViT-B/16 backbone pre-trained with DINO. The output of \verb+[CLS]+ tokens is utilized as the feature of the image. For the training details, the AdaptGCD is trained with a batch size of 128 for 200 epochs with an initial learning rate of 0.1 decayed with a cosine schedule on each dataset. For the adapter configuration, we utilize 8 experts with the bottleneck dimension $\hat{d}$ set as 64. 
\begin{table}
\begin{center}
\renewcommand{\arraystretch}{1.15}
\setlength{\tabcolsep}{1pt}
\caption{Statistic information of the 7 datasets.}
\label{table:dataset_setting}
\scalebox{0.95}{
\begin{tabular}{ccccccc}
\toprule
\multirow{2}{*}{\textbf{Benchmark}} & \multirow{2}{*}{\textbf{Datasets}} & \multirow{2}{*}{\textbf{Balance}} & \multicolumn{2}{c}{\textbf{Labeled}} & \multicolumn{2}{c}{\textbf{Unlabeled}}\\
\cmidrule(r){4-5} \cmidrule(r){6-7} 
 & & & \textbf{\# Image} & \textbf{\# Class} & \textbf{\# Image} & \textbf{\# Class}\\
\midrule
\multirow{3}{*}{\shortstack{Semantic Shift\\Benchmark}} & CUB-200 \cite{welinder2010caltech} & \Checkmark & 1.5K & 100 & 4.5K & 200 \\
&Aircraft \cite{DBLP:journals/corr/MajiRKBV13} & \Checkmark & 1.7K & 50 & 5.0K & 100 \\
&Stanford Cars \cite{krause20133d}    & \Checkmark & 2.0K & 98 & 6.1K & 196 \\
\midrule
{\multirow{3}{*}{\shortstack{{Generic Image}\\ {Recognition}\\Benchmark}}}  &CIFAR10 \cite{krizhevsky2009learning} & \Checkmark & 12.5K & 5 & 37.5K & 10 \\
&CIFAR100 \cite{krizhevsky2009learning} & \Checkmark & 20.0K & 80 & 30.0K & 100 \\
&ImageNet100 \cite{DBLP:conf/eccv/TianKI20} & \Checkmark & 31.9K & 50 & 95.3K & 100 \\
\midrule
{\shortstack{Long-tailed}}&Herbarium19 \cite{tan2019herbarium} & \XSolidBrush & 8.9K & 341 & 25.4K & 683 \\
\bottomrule
\end{tabular}
}
\end{center}
\end{table}
\begin{table}[!htbp]
\begin{center}
\renewcommand{\arraystretch}{1.1}
\setlength{\tabcolsep}{3.5pt}
\caption{Detailed hyperparamter configuration of AdaptGCD on 7 datasets}
\label{table:hyper_setting_adapter}
\scalebox{1}{
\begin{tabular}{cccccccccc}
\hline\noalign{\smallskip}
\multirow{2}{*}{\textbf{Benchmark}} & \multirow{2}{*}{\textbf{Datasets}} &  \multicolumn{8}{c}{\textbf{Configuration}}\\
\cmidrule{3-10}
& & $s$ & $P$ & $T$ & $\hat{d}$ & $\alpha$ & $\beta$ & $\tau_r$ & $\tau_g$ \\
\noalign{\smallskip}
\hline
\noalign{\smallskip}
\multirow{3}{*}{\shortstack{Semantic Shift\\Benchmark}}&CUB-200 \cite{welinder2010caltech} & 0.4 & 6 & 8 & 64 & 0.03 & 0.1 & 5 & 0.1\\
&Aircraft \cite{DBLP:journals/corr/MajiRKBV13} & 0.4 & 8 & 8 & 64 & 0.1 & 0.1 & 10 & 0.1 \\
&Scars \cite{krause20133d} & 0.4 & 6 & 8 & 64 & 0.1 &  0.1 & 10 & 0.1  \\
\midrule
\multirow{3}{*}{\shortstack{Generic Image\\Recognition\\Benchmark}}&CIFAR10 \cite{krizhevsky2009learning} & 0.2 & 6 & 8 & 64 & 0.05 & 0.05 & 10 & 0.1   \\
&CIFAR100 \cite{krizhevsky2009learning} & 0.8 & 8 & 8 & 64 & 0.05 &  0.05 & 10 & 0.1  \\
&ImageNet100 \cite{DBLP:conf/eccv/TianKI20} & 0.4 & 8 & 8 & 64 & 0.06 &  0.2 & 10 & 0.1  \\
\midrule
{\shortstack{Long-tailed}}&Herbarium19 \cite{tan2019herbarium} & 0.4 & 8 & 8 & 64 & 0.05   &  0.05 & 10 & 0.1  \\
\noalign{\smallskip}
\hline
\end{tabular}
}
\end{center}
\end{table}
The number of adapted blocks $P$ is assigned to be 6 on CUB-200, Scars and CIFAR10. As for the Aircraft, CIFAR100, Herbarium19 and Imagenet100, which present greater challenges for the classification, $P$ is designated to be 8. {With these hyper-parameters, the proposed method is equipped with few trainable parameters in the backbone, totaling less than 6.4M, which is lower than the 7M in the last block trained in SimGCD, thus ensuring fairness. More analysis about the number of tunable parameters is presented on Appendix.} For expert assignments, 4 experts are utilized as old-class experts while another 4 experts are selected as new-class experts, i.e. $|\mathcal{T}^{old}|=|\mathcal{T}^{new}|=4$. Further, Table \ref{table:hyper_setting_adapter} shows the rest hyper-parameter settings for the 7 datasets. {For the integration of the proposed method and SPTNet, we follow the basic hyperparameter settings of both methods and conduct experiments on 5 datasets due to limited computational resources. More specific settings of hybrid methods are also detailed in the Appendix.}

\emph{3) Evaluation:}
The performance is evaluated with the clustering accuracy for old, new and all classes as \cite{DBLP:conf/cvpr/gcdVazeHVZ22}. Given the ground truth $y_i$ and the predicted labels $\hat{y}_i$ for each $x_i$, the accuracies for the old/new/all classes $acc^{old}/acc^{new}/acc^{all}$ are formulated as Eq. (\ref{cal::metrics}).
\begin{equation}
\label{cal::metrics}
    \begin{aligned}
        acc^{old}=& \frac{1}{|\mathcal{D}^{u,old}|}\sum_{i=1}^{|\mathcal{D}^{u,old}|}\mathbbm{1}(y_i={\rm Purmute}(\hat{y}_i))\\
        acc^{new}=& \frac{1}{|\mathcal{D}^{u,new}|}\sum_{i=1}^{|\mathcal{D}^{u,new}|}\mathbbm{1}(y_i={\rm Purmute}(\hat{y}_i))\\
        acc^{all}=& \frac{1}{|\mathcal{D}^u|}\sum_{i=1}^{|\mathcal{D}^u|}\mathbbm{1}(y_i={\rm Permute}(\hat{y}_i))
    \end{aligned}
\end{equation}
where the $\mathcal{D}^{u,old}$ and $\mathcal{D}^{u,new}$ are the subsets of  $\mathcal{D}^{u}$ only containing old and new classes. $\rm Permute(\cdot)$ is the optimal permutation matching the predicted cluster assignments to the ground truth class labels.

\begin{table*}[!htbp]
    \renewcommand{\arraystretch}{1.15}
    \setlength{\tabcolsep}{10.6pt}
    \caption{Evaluation results on the generic image recognition benchmark. Bold values represent the best results, while underlined values denote the second best results.}
    \label{table:generic}
    \centering
    \scalebox{1}{
    \begin{threeparttable}
    \begin{tabular}{ccccccccccc}
    \toprule
    \multirow{3}{*}{\textbf{Methods}} & \multirow{3}{*}{\textbf{Tuning Strategy}} & \multicolumn{9}{c}{ Generic Image Recognition Benchmark}\\
    \cmidrule(lr){3-11}
    & & \multicolumn{3}{c}{CIFAR10} & \multicolumn{3}{c}{CIFAR100} & \multicolumn{3}{c}{ImageNet100}\\
    \cmidrule(lr){3-5}\cmidrule(lr){6-8}\cmidrule(lr){9-11}
     & & \cellcolor{blue!10} All & Old & New & \cellcolor{blue!10} All & Old & New & \cellcolor{blue!10} All & Old & New \\
    \midrule
    $k\mbox{-}means$ \cite{DBLP:conf/soda/ArthurV07}                &    -         &\cellcolor{blue!10}83.6&85.7&82.5&\cellcolor{blue!10}52.0&52.2&50.8&\cellcolor{blue!10}72.7&75.5&71.3\\ 
    RankStats+ \cite{DBLP:journals/pami/HanREVZ22}            &   Partial    &\cellcolor{blue!10}46.8&19.2&60.5&\cellcolor{blue!10}58.2&77.6&19.3&\cellcolor{blue!10}37.1&61.6&24.8\\ 
    UNO+ \cite{DBLP:conf/iccv/FiniSLZN021}                    &   Partial    &\cellcolor{blue!10}68.6&\textbf{98.3}&53.8&\cellcolor{blue!10}69.5&80.6&47.2&\cellcolor{blue!10}70.3&95.0&57.9\\  
    \midrule
    ORCA \cite{DBLP:conf/iclr/CaoBL22}                        &   Partial    &\cellcolor{blue!10}81.8&86.2&79.6&\cellcolor{blue!10}69.0&77.4&52.0&\cellcolor{blue!10}73.5&92.6&63.9\\ 
    GCD \cite{DBLP:conf/cvpr/gcdVazeHVZ22}                    &   Partial    &\cellcolor{blue!10}91.5&\underline{97.9}&88.2&\cellcolor{blue!10}73.0&76.2&66.5&\cellcolor{blue!10}74.1&89.8&66.3\\
    DCCL \cite{DBLP:conf/cvpr/PuZS23}                         &   Partial    &\cellcolor{blue!10}96.3&96.5&96.9&\cellcolor{blue!10}75.3&76.8&70.2&\cellcolor{blue!10}80.5&90.5&76.2\\ 
    SimGCD \cite{DBLP:conf/iccv/simgcdWenZQ23}                &   Partial    &\cellcolor{blue!10}97.1&95.1&98.1&\cellcolor{blue!10}80.1&81.2&{77.8}&\cellcolor{blue!10}83.0&93.1&77.9\\ 
    CMS \cite{DBLP:conf/cvpr/ChoiKC24}                        &   Partial    &-&-&-&\cellcolor{blue!10}{82.3}&85.7&75.5&\cellcolor{blue!10}84.7&\textbf{95.6}&79.2\\ 
    AMEND \cite{DBLP:conf/wacv/BanerjeeKB24}& Partial & \cellcolor{blue!10}96.8&94.6&97.8&\cellcolor{blue!10}81.0&79.9&\textbf{83.3}&\cellcolor{blue!10}83.2&92.9&78.3\\
    
    PromptCAL \cite{DBLP:conf/cvpr/promptcalgcdZhangKSNCK23}  &   Prompt     &\cellcolor{blue!10}\textbf{97.9}&{96.6}&98.5&\cellcolor{blue!10}81.2&84.2&75.3&\cellcolor{blue!10}83.1&92.7&78.3\\ 
    GCA+PromptCAL \cite{DBLP:conf/wacv/OtholtMY24} & Prompt & \cellcolor{blue!10}95.5&95.9&95.2&\cellcolor{blue!10}82.4&85.6&75.9&\cellcolor{blue!10}82.8&94.1&77.1 \\
    SPTNet \cite{wang2024sptnet}                              &   Prompt     &\cellcolor{blue!10}{97.3}&95.0&\underline{98.6}&\cellcolor{blue!10}{81.3}&{84.3}&75.6&\cellcolor{blue!10}\underline{85.4}&{93.2}&\underline{81.4}\\   
    \midrule
    \rowcolor{blue!10} AdaptGCD                                     &   Adapter    &\textbf{97.9}&95.5&\textbf{99.0}&\textbf{84.0}& \textbf{86.2}&\underline{79.7}& \textbf{86.4}&\underline{94.9}&\textbf{82.1}\\ 
    \rowcolor{blue!10}AdaptGCD+SPTNet                               &   Adapter    &\underline{97.7}&95.2&\textbf{99.0}&\underline{83.0}&\underline{86.1}&76.8&-&-&-\\  
    \bottomrule
    \end{tabular}
    \end{threeparttable}
    }
\end{table*}



\begin{table*}[!htbp]
\renewcommand{\arraystretch}{1.1}
\setlength{\tabcolsep}{7pt}
\caption{Evaluation results on the semantic shift benchmark and long-tailed dataset. Bold values represent the best results, while underlined values denote the second best results.}
\label{table:ssb}
\centering
\scalebox{1}{
\begin{threeparttable}
\begin{tabular}{cccccccccccccc}
\toprule
\multirow{3}{*}{\textbf{Methods}} & \multirow{3}{*}{\textbf{Tuning Strategy}} &\multicolumn{9}{c}{ Semantic Shift Benchmark} & \multicolumn{3}{c}{Long-tailed Dataset}\\
\cmidrule(lr){3-11}\cmidrule(lr){12-14}
& & \multicolumn{3}{c}{CUB-200} & \multicolumn{3}{c}{Scars} & \multicolumn{3}{c}{Aircraft} & \multicolumn{3}{c}{Herbarium19}\\
\cmidrule(lr){3-5}\cmidrule(lr){6-8}\cmidrule(lr){9-11}\cmidrule(lr){12-14}
& & \cellcolor{blue!10}All & Old & New & \cellcolor{blue!10}All & Old & New & \cellcolor{blue!10}All & Old & New  & \cellcolor{blue!10}All & Old & New \\
\midrule
$k\mbox{-}means$ \cite{DBLP:conf/soda/ArthurV07}                &    -         &\cellcolor{blue!10}34.3&38.9&32.1&\cellcolor{blue!10}12.8&10.6&13.8&\cellcolor{blue!10}16.0&14.4&16.8&\cellcolor{blue!10}13.0&12.2&13.4\\
RankStats+ \cite{DBLP:journals/pami/HanREVZ22}            &   Partial    &\cellcolor{blue!10}33.3&51.6&24.2&\cellcolor{blue!10}28.3&61.8&12.1&\cellcolor{blue!10}26.9&36.4&22.2&\cellcolor{blue!10}27.9&55.8&12.8\\
UNO+ \cite{DBLP:conf/iccv/FiniSLZN021}                    &   Partial    &\cellcolor{blue!10}35.1&49.0&28.1&\cellcolor{blue!10}35.5&70.5&18.6&\cellcolor{blue!10}40.3&56.4&32.2&\cellcolor{blue!10}28.3&53.7&14.7\\
\midrule
ORCA \cite{DBLP:conf/iclr/CaoBL22}                        &   Partial    &\cellcolor{blue!10}35.3&45.6&30.2&\cellcolor{blue!10}23.5&50.1&10.7&\cellcolor{blue!10}22.0&31.8&17.1&\cellcolor{blue!10}35.4&51.0&27.0\\
GCD \cite{DBLP:conf/cvpr/gcdVazeHVZ22}                    &   Partial    &\cellcolor{blue!10}51.3&56.6&48.7&\cellcolor{blue!10}39.0&57.6&29.9&\cellcolor{blue!10}45.0&41.1&46.9&\cellcolor{blue!10}20.9&30.9&15.5\\
DCCL \cite{DBLP:conf/cvpr/PuZS23}                         &   Partial    &\cellcolor{blue!10}63.5&60.8&64.9&\cellcolor{blue!10}43.1&55.7&36.2&-&-&-&-&-&-\\
SimGCD \cite{DBLP:conf/iccv/simgcdWenZQ23}                &   Partial    &\cellcolor{blue!10}60.3&65.6&57.7&\cellcolor{blue!10}53.8&71.9&45.0&\cellcolor{blue!10}54.2&59.1&51.8&\cellcolor{blue!10}44.0&58.0&\underline{36.4}\\
CMS \cite{DBLP:conf/cvpr/ChoiKC24}                        &   Partial    &\cellcolor{blue!10}{68.2}&\textbf{76.5}&{64.0}&\cellcolor{blue!10}56.9&76.1&47.6&\cellcolor{blue!10}56.0&63.4&52.3&\cellcolor{blue!10}36.4&54.9&26.4\\
AMEND \cite{DBLP:conf/wacv/BanerjeeKB24}& Partial &\cellcolor{blue!10}64.9&75.6&59.6&\cellcolor{blue!10}56.4&73.3&48.2&\cellcolor{blue!10}52.8&61.8&48.3&\cellcolor{blue!10}\underline{44.2}&\underline{60.5}&{35.4}\\
$\mu$GCD \cite{DBLP:conf/nips/VazeVZ23}& Partial &\cellcolor{blue!10}65.7&68.0&64.6&\cellcolor{blue!10}56.5&68.1&50.9&\cellcolor{blue!10}53.8&55.4&53.0& - & - &-\\
PromptCAL \cite{DBLP:conf/cvpr/promptcalgcdZhangKSNCK23}  &   Prompt     &\cellcolor{blue!10}62.9&64.4&62.1&\cellcolor{blue!10}50.2&70.1&40.6&\cellcolor{blue!10}52.2&52.2&52.3&\cellcolor{blue!10}37.0&52.0&28.9\\
GCA+PromptCAL \cite{DBLP:conf/wacv/OtholtMY24} & Prompt & \cellcolor{blue!10}\underline{68.8}&73.4&\underline{66.6}&\cellcolor{blue!10}54.4&72.1&45.8&\cellcolor{blue!10}52.0&57.1&49.5& - & - &-\\
SPTNet \cite{wang2024sptnet}                              &   Prompt     &\cellcolor{blue!10}{65.8}&{68.8}&{65.1}&\cellcolor{blue!10}{59.0}&\underline{79.2}&{49.3}&\cellcolor{blue!10}\underline{59.3}&{61.8}&\textbf{58.1}&\cellcolor{blue!10}{43.4}&{58.7}&{35.2}\\
\midrule
\rowcolor{blue!10} AdaptGCD                                     &   Adapter    &\underline{68.8}&{74.5}&{65.9}&\underline{62.7}&\textbf{80.6}&\underline{54.0}&{57.9}&\underline{65.2}&{54.3}&\textbf{45.7}&\textbf{60.6}&\textbf{37.7}\\
\rowcolor{blue!10}AdaptGCD+SPTNet                               &   Adapter    &\cellcolor{blue!10}\textbf{71.6}&\underline{75.7}& \textbf{69.6}&\cellcolor{blue!10}\textbf{63.5}&75.7&\textbf{57.6}&\cellcolor{blue!10}\textbf{60.8}&\textbf{68.1}&\underline{57.2}&-&-&-\\
\bottomrule
\end{tabular}
\end{threeparttable}
}
\end{table*}

\subsection{Comparison with State-of-the-Arts}
\emph{1) Evaluation on Generic Image Recognition Datasets:} AdaptGCD is compared with the previous advanced methods, including the methods based on partial tuning (ORCA \cite{DBLP:conf/iclr/CaoBL22}, GCD \cite{DBLP:conf/cvpr/gcdVazeHVZ22}, SimGCD \cite{DBLP:conf/iccv/simgcdWenZQ23}, DCCL \cite{DBLP:conf/cvpr/PuZS23}, CMS \cite{DBLP:conf/cvpr/ChoiKC24}, AMEND \cite{DBLP:conf/wacv/BanerjeeKB24}, etc.) and those based on prompt tuning (PromptCAL \cite{DBLP:conf/cvpr/promptcalgcdZhangKSNCK23}, GCA+PromptCAL \cite{DBLP:conf/wacv/OtholtMY24} and SPTNet \cite{wang2024sptnet}). The results on the generic image recognition benchmark are shown in Table {\ref{table:generic}}. It is observed that our AdaptGCD manifests superior performance across three general benchmarks. Compared to the baseline SimGCD, the AdaptGCD, which is equipped with fewer tunable parameters, achieves improvements of 0.8\%, 3.9\%, and 3.4\% for ``All'' classes on three datasets, demonstrating the effectiveness of adapter tuning in striking the balance between the generalization and adaptability. When juxtaposed with other concurrent methods, our method advances on almost all the metrics. 
On CIFAR10 datasets, AdaptGCD prioritizes a balance between the old and new categories so that it outperforms the GCD by 10.8\% in ``New'' classes, at a cost of 2.4\% dip in the ``Old'' classes. Note that the most advanced prompt-based method SPTNet underperforms its baseline SimGCD on several metrics on CIFAR10 and CIFAR100 datasets since the low-resolution images hinder the utility of spatial prompts. However, poor image quality exerts a minimal impact on the proposed method. The promising performance achieved by our AdaptGCD validates that the adapter-like methods could exhibit excellent adaptability to data with varying image qualities than the prompt-like rivals. Furthermore, as a plug-in module, it also improves the performance of the existing method SPTNet.

\emph{2) Evaluation on Fine-Grained Datasets:} Table {\ref{table:ssb}} presents the results on fine-grained datasets including the semantic shift benchmark and the long-tailed dataset Herbarium19. Due to the large intra-class and small inter-class variations for data, it is more difficult to discover fine-grained categories than the generic classes. Despite the difficulty, AdaptGCD shows consistently superior performance on the CUB-200, Scars and Herbarium19 datasets, achieving an average improvement of 3\% on ``All'' classes in comparison with the advanced results from SPTNet due to the excellent adaptability of adapter tuning. Even within the Aircraft dataset, it achieves the second best scores with only 1.4\% lag on ``All'' classes. These observations indicate the efficacy of the proposed method across fine-grained datasets. Additionally, after incorporating the SPTNet, the proposal further achieves better performance on the 'All' classes across the three datasets. This highlights the scalability of the proposed module and its potential to function as a plug-and-play component.

\begin{table*}[!htbp]
    \begin{center}
    \renewcommand{\arraystretch}{1.5}
    \setlength{\tabcolsep}{2.5pt}
    \caption{Ablation study on the semantic shift benchmark and generic image recognition benchmark. $T$ is the number of experts while $\hat{d}$ indicates the dimension number of each expert.} 
    \label{table::abalation}
    \begin{tabular}{clccccccccccccccccccccc}
    \toprule
    \multirow{3}{*}{\textbf{Index}}&\multirow{3}{*}{\textbf{Method}} & \multirow{3}{*}{$T$}  & \multirow{3}{*}{$\hat{d}$} &  \multirow{3}{*}{\textbf{Loss}} & \multicolumn{9}{c}{Semantic Shift Benchmark} & \multicolumn{9}{c}{Generic Image Recognition Benchmark}\\
    \cmidrule(r){6-14} \cmidrule(r){15-23}
    & & & & & \multicolumn{3}{c}{CUB-200} & \multicolumn{3}{c}{Scars} & \multicolumn{3}{c}{Aircraft}& \multicolumn{3}{c}{CIFAR100} & \multicolumn{3}{c}{CIFAR10} & \multicolumn{3}{c}{Imagenet100}\\
    \cmidrule(lr){6-8}\cmidrule(lr){9-11}\cmidrule(lr){12-14}\cmidrule(lr){15-17}\cmidrule(lr){18-20}\cmidrule(lr){21-23}
    & & & & & \cellcolor{blue!10}All & Old & New & \cellcolor{blue!10}All & Old & New & \cellcolor{blue!10}All & Old & New & \cellcolor{blue!10}All & Old & New & \cellcolor{blue!10}All & Old & New & \cellcolor{blue!10}All & Old & New\\
    \midrule
    {\tt (1)} & SimGCD & - & - & $\mathcal{L}_{sgcd}$ &\cellcolor{blue!10}60.3 &65.6 &57.7 &\cellcolor{blue!10}53.8 &71.9 &45.0& \cellcolor{blue!10}54.2& 59.1& 51.8 &  \cellcolor{blue!10}80.1 & 81.2 & \underline{77.8} & \cellcolor{blue!10}97.1 & 95.1 & 98.1 & \cellcolor{blue!10}83.0 & 93.1 & 77.9  \\
    \midrule
    {\tt (2)}&{AdaptGCD} & 1 & 64	    &$\mathcal{L}_{sgcd}$ & \cellcolor{blue!10}67.5 & \underline{75.3} & 63.5&\cellcolor{blue!10}59.8	&77.4	&51.2&\cellcolor{blue!10}55.2	&65.0	&50.3 & \cellcolor{blue!10}80.7 & {85.6} & 70.7 & \cellcolor{blue!10}\textbf{97.9} & \textbf{96.0} & 98.9 & \cellcolor{blue!10}84.6 & \textbf{95.1} & 79.3 \\
    {\tt (3)}&{AdaptGCD} & 1 &512	&$\mathcal{L}_{sgcd}$ & \cellcolor{blue!10}68.0	&\textbf{76.1}	&63.9	&\cellcolor{blue!10}61.3	&77.5	&\underline{53.5}&\cellcolor{blue!10}\underline{57.7}	&\textbf{66.9}	&53.1	&\cellcolor{blue!10}81.9&\textbf{87.9}&69.7 & \cellcolor{blue!10}\textbf{97.9}&\underline{95.8}&\underline{99.0} &\cellcolor{blue!10}85.0&\textbf{95.1}&80.0 \\
    \midrule
    \rowcolor{blue!10}{\tt (4)}&AdaptGCD   & 8& 64 & $\mathcal{L}_{sgcd}$&\cellcolor{blue!10}68.5 &	74.9 &	{65.3}&\cellcolor{blue!10}61.4 &	79.4 &	52.7 &	\cellcolor{blue!10}57.3 &	\underline{65.3} &	53.2 & \cellcolor{blue!10}82.6 & 87.1 & 73.6 & \cellcolor{blue!10}\underline{97.8} & 95.2 & \textbf{99.1} & \cellcolor{blue!10}85.4 & \textbf{95.1} & 80.6 \\
    \rowcolor{blue!10}{\tt (5)}&AdaptGCD   & 8& 64 & $\mathcal{L}_{sgcd}+\mathcal{L}_{ba}$&\cellcolor{blue!10}\underline{68.6} &	75.0 &	\underline{65.4}&\cellcolor{blue!10}\underline{62.0} &	\textbf{80.9} &	52.8 &	\cellcolor{blue!10}\underline{57.7} &	\underline{65.3} &	\underline{53.9} & \cellcolor{blue!10}\underline{83.5} & \underline{87.2} & 76.0 & \cellcolor{blue!10}\underline{97.8} & 95.3 & \textbf{99.1} & \cellcolor{blue!10}\underline{86.0} & \textbf{95.1} & \underline{81.4} \\
     \rowcolor{blue!10}{\tt (6)}&AdaptGCD  & 8& 64 &$\mathcal{L}_{sgcd}+\mathcal{L}_{ba}+\mathcal{L}_{cba}$ &\textbf{68.8}&{74.5}&\textbf{65.9}&\textbf{62.7}&\underline{80.6}&\textbf{54.0}&\textbf{57.9}&{65.2}&\textbf{54.3}&\textbf{84.0} & 86.2 & \textbf{79.7} & \textbf{97.9} & 95.5 & \underline{99.0} & \textbf{86.4} & \underline{94.9} & \textbf{82.1} \\
    \bottomrule
    \end{tabular}
    \end{center}
\end{table*}

\subsection{Ablation Study}

To investigate the impact of different modules, we conduct extensive ablation studies on the semantic shift benchmark and generic image recognition benchmark following \cite{wang2024sptnet,DBLP:conf/cvpr/promptcalgcdZhangKSNCK23}. The results are reported in Table \ref{table::abalation}.

\emph{1) Influence of Adding Adapter Structure:} Firstly, we have validated the role of a single adapter as illustrated in {\tt (2)}. Given that the bottleneck dimension is set as 64, the count of learnable parameters in the backbone is less than 0.8M, which is significantly less than the number of parameters in the last block (approximately 7M). Despite the less trainable parameters in the backbone, the adapter-based method still achieves a remarkable accuracy boost of 7.2\%, 6.0\%, 1.0\%, 0.6\%, 0.8\% and 1.6\% for ``All" classes on the CUB-200, Scars, Aircraft, CIFAR100, CIFAR10 and Imagenet100 datasets. These findings affirm the adapter's efficacy in GCD tasks in preserving the general knowledge and enhancing the adaptability to GCD task. 

\emph{2) Influence of Multi-Expert Adapter and Route Assignment Constraint:} To validate the effectiveness of the multi-expert adapter (MEA), the method implemented with MEA (in {\tt(4)}) is compared with two alternate single adapter approaches. One incorporates an equivalent bottleneck dimension, as shown in {\tt(2)}, while the other uses the setup with a comparable parameter count, denoted in {\tt(3)}. With an identical bottleneck dimension, the results reveal that the MEA outperforms the single expert method, thereby establishing the MEA's superiority. On the other hand, when the number of learnable parameters is similar (see {\tt(3)} vs. {\tt(4)}), the MEA does not show obvious advantages. Nevertheless, with the assistance of route assignment constraints (see {\tt(5),(6)}), the methods incorporating MEA surpass the single adapter method {\tt(3)}, in terms of ``All" classes across six datasets. This statement emphasizes the significant role that the route assignment constraint holds in the MEA training process. Actually, on average across three fine-grained and three general datasets, the two assignment losses $\mathcal{L}_{cba}$ and $\mathcal{L}_{ba}$ improve all class accuracy by 0.7\% and 0.8\%. It is further demonstrated that the improvements derived from the application of MEA are not only due to the increased number of learnable parameters but also contribute to the separation of different data via route allocations, enabled by the constraint. Moreover, the two losses enhance the performance of ``New'' classes by an average of 1\% and 2.5\% across fine-grained and general datasets, albeit at the expense of a slight reduction in ``Old'' classes (see {\tt(6)} vs. {\tt(4)}). This observation further validates our motivation to alleviate the bias towards old classes. 
More analysis about MEA is presented in the discussion section.

\begin{figure*}[!htbp]
\center
\includegraphics[width=0.32\textwidth]{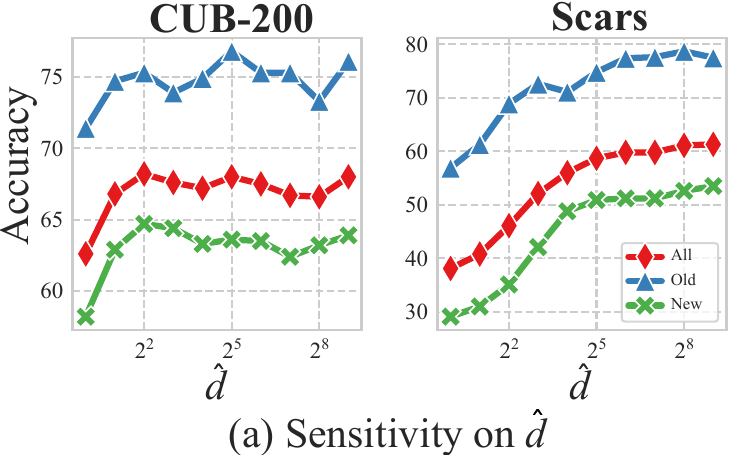}
\includegraphics[width=0.32\textwidth]{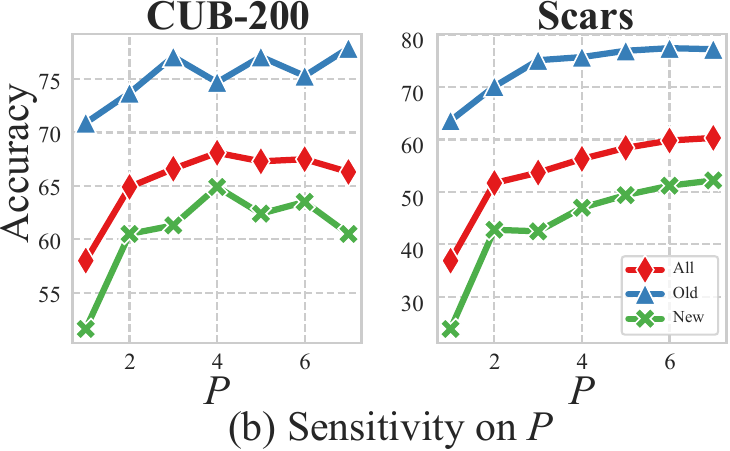}
\includegraphics[width=0.32\textwidth]{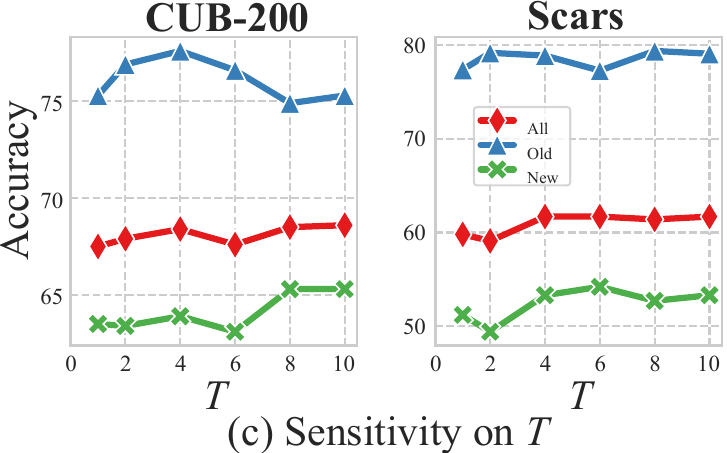}
\caption{{Sensitivity analysis on three critical hyperparameters.}  (a) bottleneck dimension $\hat{d}$, (b) the number of adapted blocks $P$, (c) expert count $T$.}
\label{fig::layers} 
\end{figure*}

\subsection{Sensitivity Analysis}

\emph{1) Bottleneck Dimension, Number of Adapter Blocks and Expert Count:}
We evaluated the impact of three key hyper-parameters of the adapter: bottleneck dimension $\hat{d}$, number of adapted blocks $P$ and expert count $T$, and show the results in Fig. \ref{fig::layers}. First, we conduct experiments with models employing single adapters across different $\hat{d}$ and report the results in Fig. \ref{fig::layers} (a). Remarkably, when the bottleneck dimension is relatively small, there is a noticeable rise in performance with the increase in the bottleneck dimension. As the dimension further elevates, performance fluctuates within a certain range due to overfitting caused by the excessive amount of parameters. A similar trend is also observed as the increase in the number of adapted blocks $P$ in Fig. \ref{fig::layers} (b), i.e., as $P$ is large enough, inserting more adapters in the blocks yields additional resources but limited performance gain. It also validates the effectiveness of only inserting adapters in parts of blocks. For the high performance of a single adapter, the impact of $T$ in Fig. \ref{fig::layers} (c) is not so noticeable, but the increment in expert count still results in a performance boost for ``All'' classes, further corroborating the effectiveness of multi-expert adapters.

\emph{2) Expert Assignment:}
To verify the impact of expert distribution, we conducted experiments with different numbers of new and old experts on the Scars and Aircraft datasets, with results shown in Table \ref{tab:sensity_expertass}. It is observed that the model's accuracy in ``All'' classes presents a unimodal distribution with respect to the number of old-class experts. On the two dataset, the performance is highest when the number of new and old experts is equal. This indicates that overemphasizing either new-class or old-class experts negatively affects performance due to the disruption of balance. Additionally, we noted that in both datasets, when the number of old-class experts exceeds that of new-class experts, the accuracy in the ``Old'' classes remains nearly unchanged or increases, while the accuracy in the ``New'' classes significantly declines, and vice versa. This aligns with the motivation behind our MEA module.

    \begin{table}[tp]
    \caption{Performance of the proposed AdaptGCD equipped with different number of old and new experts on Aircraft and Scars datasets.}
    \renewcommand{\arraystretch}{1.2}
    \label{tab:sensity_expertass}
    \centering
        \scalebox{1}
        {\begin{tabular}{cccccccc}
            \toprule
            \multirow{2}{*}{$|\mathcal{T}^{old}|$} & \multirow{2}{*}{$|\mathcal{T}^{new}|$}  & \multicolumn{3}{c}{Aircraft} & \multicolumn{3}{c}{Scars} \\
            \cmidrule(r){3-5} \cmidrule(r){6-8}
             & & \cellcolor{blue!10}All & Old & New & \cellcolor{blue!10}All & Old & New\\
            \midrule
            2  &  6	 &\cellcolor{blue!10}57.0  &	63.6 &	53.7 &\cellcolor{blue!10} 61.7 & 76.6 & \textbf{54.6} \\
            3  &  5	 &\cellcolor{blue!10}57.4  &	63.9 &	54.2 &\cellcolor{blue!10} 62.4 & 79.7 & 54.0 \\
            \rowcolor{blue!10}4  &  4	 &
            \textbf{57.9}  & \textbf{65.2}   & \textbf{54.3} & 
   \textbf{ 62.7}& 80.6 & 54.0 \\ 
            5  &  3	 &\cellcolor{blue!10}57.4  &	64.8 &	53.7 &\cellcolor{blue!10} 62.0 & 79.6 & 53.5 \\
            6  &  2	 &\cellcolor{blue!10}55.9  &	65.1 &	51.3 &\cellcolor{blue!10} 61.7 & \textbf{81.1} & 52.3 \\
            \bottomrule
            \end{tabular}}
            
\end{table}

\subsection{Discussion}
\label{sec::discussion}

\emph{1) Scalability with DINOv2 Backbone:} To validate the scalability of the proposal across different pretrained models, we conduct experiments adapting our AdaptGCD to the DINOv2 \cite{DBLP:journals/tmlr/OquabDMVSKFHMEA24} backbone, as shown in Table \ref{table::dinov2}. Compared to the advanced method, our proposal achieves an overall performance improvement on the ``All'' classes across three datasets. 
These observations further illustrate the high adaptability of the proposed adapter-based approach for different pre-trained models.
\begin{table}
\begin{center}
\renewcommand{\arraystretch}{1.3}
\setlength{\tabcolsep}{3.45pt}
\caption{Performance on the semantic shift benchmark with the pretrained DINOv2 model.}
\label{table::dinov2}
\scalebox{0.9}{
\begin{tabular}{lcccccccccc}
\toprule
\multirow{2}{*}{\textbf{Methods}}& \multirow{2}{*}{\textbf{Backbone}} & \multicolumn{3}{c}{CUB-200} & \multicolumn{3}{c}{Scars} & \multicolumn{3}{c}{Aircraft}\\
\cmidrule(lr){3-5}\cmidrule(lr){6-8}\cmidrule(lr){9-11}
 &  & \cellcolor{blue!10} All & Old & New & \cellcolor{blue!10} All & Old & New & \cellcolor{blue!10} All & Old & New \\
\midrule
$k$-means \cite{DBLP:conf/soda/ArthurV07} &  DINOv2  & \cellcolor{blue!10}67.6 & 60.6 & 71.1 & \cellcolor{blue!10}29.4 & 24.5 & 31.8 & \cellcolor{blue!10}18.9 & 16.9 & 19.9 \\
GCD     \cite{DBLP:conf/cvpr/gcdVazeHVZ22}   &  DINOv2  & \cellcolor{blue!10}71.9 & 71.2 & 72.3 & \cellcolor{blue!10}65.7 & 67.8 & 64.7 & \cellcolor{blue!10}55.4 & 47.9 & 59.2 \\
SimGCD  \cite{DBLP:conf/iccv/simgcdWenZQ23}   &  DINOv2  & \cellcolor{blue!10}71.5 & \underline{78.1} & 68.3 & \cellcolor{blue!10}71.5 & 81.9 & {66.6} & \cellcolor{blue!10}63.9 & \underline{69.9} & 60.9 \\
$\mu$GCD \cite{DBLP:conf/nips/VazeVZ23}  &  DINOv2  & \cellcolor{blue!10}74.0 & 75.9 & 73.1 & \cellcolor{blue!10}\textbf{76.1} & \textbf{91.0} & \underline{68.9} & \cellcolor{blue!10}\underline{66.3} & 68.7 & \textbf{65.1} \\
CiPR  \cite{DBLP:journals/tmlr/Hao0W24}     &  DINOv2  & \cellcolor{blue!10}\underline{78.3} & 73.4 & \textbf{80.8} & \cellcolor{blue!10}66.7 & 77.0 & 61.8 & \cellcolor{blue!10}- & - & - \\
\midrule
\rowcolor{blue!10} AdaptGCD  & DINOv2 & \cellcolor{blue!10}\textbf{78.7} & \textbf{82.1} & \underline{77.0} & \cellcolor{blue!10}\underline{75.8} & \underline{85.5} & \textbf{71.1} & 
\cellcolor{blue!10}\textbf{67.7} & \textbf{73.3} & \underline{64.9}
\\
\bottomrule
\end{tabular}
}
\end{center}
\end{table}

\begin{table}
\begin{center}
\renewcommand{\arraystretch}{1.3}
\setlength{\tabcolsep}{4pt}
\caption{Performance of different PEFT methods on CUB-200 and Scars datasets. \textbf{\#Params} indicates the number of tunable parameters in backbone of the method. }
\label{table:lora_lst_}
\scalebox{0.95}{
\begin{tabular}{cccccccccccc}
\toprule
\multirow{2}{*}{\textbf{Methods}}& 
\multirow{2}{*}{\textbf{\#Params}}& \multicolumn{3}{c}{CUB-200} & \multicolumn{3}{c}{Scars}\\
\cmidrule(lr){3-5}\cmidrule(lr){6-8}
& & \cellcolor{blue!10} All & Old & New & \cellcolor{blue!10} All & Old & New \\
\midrule
LoRA  & 0.6M & \cellcolor{blue!10}66.6 & 74.3 & 62.7 & \cellcolor{blue!10}57.6 & 73.0 & 50.2 \\
Side Tuning  & 0.7M & \cellcolor{blue!10}62.1 & 67.8 & 59.3 & \cellcolor{blue!10}51.6 & 73.3 & 41.1 \\
\midrule
\rowcolor{blue!10} AdaptGCD (single adapter) &  0.6M   & \textbf{67.5} & \textbf{75.3} & \textbf{63.5} &  \textbf{59.8} & \textbf{77.4} & \textbf{51.2} \\
\midrule
LoRA  & 4.6M & \cellcolor{blue!10}63.9 & 71.7 & 60.0 & \cellcolor{blue!10}54.5 & 72.1 & 46.0 \\
Side Tuning  & 4.4M & \cellcolor{blue!10}62.2 & 68.1 & 59.3 & \cellcolor{blue!10}53.8 & 74.9 & 43.6 \\ 
\midrule
\rowcolor{blue!10} AdaptGCD (MEA)  &  4.8M     & \textbf{68.8} & \textbf{74.5} & \textbf{65.9} &  \textbf{62.7} & \textbf{80.6} & \textbf{54.0} \\ 
\bottomrule
\end{tabular}
}
\end{center}
\end{table}

\emph{2) Comparison with  Different Efficient Tuning Methods:} To further illustrate that the performance improvement of the proposed method does not solely stem from the structure of the additional parameters, we further reproduce some other common efficient tuning methods, including LoRA \cite{DBLP:conf/iclr/HuSWALWWC22} and Side Tuning \cite{DBLP:conf/nips/Sung0B22, DBLP:conf/eccv/ZhangSZGM20} for comparison with AdaptGCD, as shown in Table \ref{table:lora_lst_}. To ensure fairness, these competitors are configured with a similar number of tunable parameters as those of the proposed single adapter and multiple adapter methods, respectively. Compared to LoRA, the proposed method incorporates nonlinear layers, introducing greater flexibility and thus achieving higher performance. In contrast to Side Tuning, 
The proposal involves slight modifications to the original Transformer rather than training an additional light Transformer, which could be more advantageous with limited labels. These observations further validate the superiority of the proposed AdaptGCD.

\begin{figure*}[!htbp]
\center
\includegraphics[width=1\textwidth]{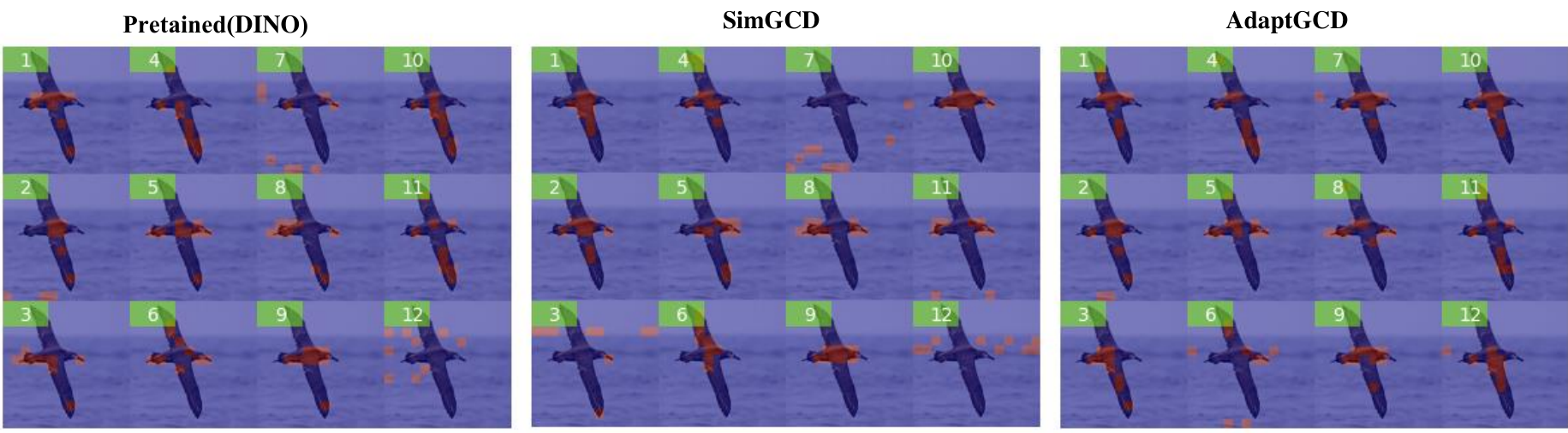}
\caption{ {Attention visualization} for 12 heads in the last blocks of backbone on CUB-200. ``Pretrained(DINO)'' represents the results from pretrained backbone without any additional training while ``SimGCD'' and ``AdaptGCD'' are those from models trained via the corresponding methods. } 
\label{fig::bird} 
\end{figure*}

\emph{3) Visualization of the Attention Maps:} To better appreciate the advantages of the AdaptGCD, we extracted the attention maps corresponding to the {\tt[CLS]} token from different heads in the last block of the ViT, highlighting the top 10 most notable patches in red for each head in Fig. \ref{fig::bird}. Upon observation, it is evident that compared to SimGCD and DINO, AdaptGCD focuses more on the foreground and exhibits more varied attention regions in different heads (in heads 3/7/11/12), showing the superior adaptability of AdaptGCD to the CUB-200 datasets. It is also worth noting that in regions like heads 3/7/11, SimGCD fails to inherit the correct attention on the foreground from the DINO model, while AdaptGCD refines it based on the pretrained model. These findings further demonstrate the motivation to protect the pretrained information.

\begin{figure}[tp]
\center
\includegraphics[width=0.48\textwidth]{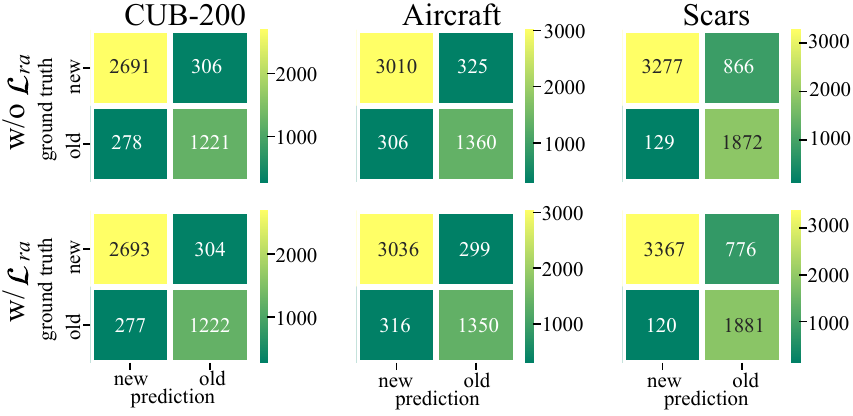}
\caption{The confusion maps for the old and new classes for the models trained with and without the route assignment losses $\mathcal{L}_{ra}$ on the semantic shift benchmark. } 
\label{fig::vis_load} 
\end{figure}

\begin{table}[!htbp]
\begin{center}
\renewcommand{\arraystretch}{1.15}
\setlength{\tabcolsep}{6pt}
\caption{Experimental results with different number of category $K$ in the network. The estimated category number is obtained using the off-the-shelf method from \cite{DBLP:conf/cvpr/gcdVazeHVZ22}. }
\label{table:estimate_}
\scalebox{0.95}{
\begin{tabular}{lccccccc}
\hline\noalign{\smallskip}
\multirow{2}{*}{\textbf{Methods}} & \multirow{2}{*}{\textbf{Known} $K$} & \multicolumn{3}{c}{CUB($K$=200)} & \multicolumn{3}{c}{Scars($K$=196)}\\
\cmidrule{3-5}\cmidrule{6-8}
 & &\cellcolor{blue!10} All & Old & New & \cellcolor{blue!10}All & Old & New\\
\noalign{\smallskip}
\hline
\noalign{\smallskip}
GCD     & \Checkmark                & \cellcolor{blue!10} 51.3 & 56.6 & 48.7 & \cellcolor{blue!10}39.0 & 57.6 & 29.9\\
SimGCD  & \Checkmark                & \cellcolor{blue!10} 60.3 & 65.6 & 57.7 & \cellcolor{blue!10}53.8 & 71.9 & 45.0\\
SPTNet & \Checkmark                & \cellcolor{blue!10} 64.6 & 70.5 & 61.6 & \cellcolor{blue!10}59.3 & 61.8 & 58.1\\
\rowcolor{blue!10}AdaptGCD& \Checkmark                &  
 { 68.8}  & 74.5   & 65.9  &62.7  & 80.6   & 54.0  
\\
\midrule
GCD     & \XSolidBrush (w/ Est.)    & \cellcolor{blue!10} 47.1 & 55.1 & 44.8 & \cellcolor{blue!10}35.0 & 56.0 & 24.8\\
SimGCD  & \XSolidBrush (w/ Est.)    & \cellcolor{blue!10} 61.5 & 66.4 & 59.1 & \cellcolor{blue!10}49.1 & 65.1 & 41.3\\
SPTNet  & \XSolidBrush (w/ Est.)    & \cellcolor{blue!10} 65.2 & 71.0 & 62.3 & \cellcolor{blue!10}- & - & - \\	
\rowcolor{blue!10}AdaptGCD& \XSolidBrush (w/ Est.)    &  \textbf{ 67.2} & \textbf{74.6} & \textbf{63.5} & \textbf{62.6} & \textbf{80.1} & \textbf{54.1}\\
\midrule
SimGCD  & \XSolidBrush (w/ 2$K$)      & \cellcolor{blue!10} 63.6 & 68.9 & 61.1 & \cellcolor{blue!10}48.2 & \textbf{64.6} & 40.2\\
\rowcolor{blue!10}AdaptGCD& \XSolidBrush (w/ 2$K$)      & \cellcolor{blue!10} \textbf{68.4} & \textbf{75.1} & \textbf{65.0} & \cellcolor{blue!10}\textbf{48.4} & {61.7} & \textbf{42.0}\\
\noalign{\smallskip}
\hline
\end{tabular}
}
\end{center}
\end{table}

\emph{4) Visualization of the Confusion Maps:} In addition, to further investigate the role of the route assignment constraint in mitigating the bias towards old classes, we also visualized the confusion matrix of methods trained with and without the $\mathcal{L}_{ra}$ on the three datasets from semantic shift benchmark in Fig. \ref{fig::vis_load}. It is observed that with the constraint $\mathcal{L}_{ra}$, the amount of new-class data misclassified as the old classes decreases, thus substantiating the effectiveness of the constraint in alleviating the preference for the old classes.

\emph{5) Scalability under Unknown Category Number:}
Due to the inaccessibility of the total number of categories (GT) in real-world scenarios, we evaluate the proposed AdaptGCD with the estimated number of categories provided by the off-the-shelf method \cite{DBLP:conf/cvpr/gcdVazeHVZ22} and a larger number of classes in Table \ref{table:estimate_}. The experiments are conducted in the CUB-200 and Scars dataset, where the ground truth of the total class number $K$ are 200 and 196 and the estimated class number $K_{est}$ given by \cite{DBLP:conf/cvpr/gcdVazeHVZ22} are 231 and 230. 
It is observed that the proposed AdaptGCD shows overall superiority under different initialized number of classes, demonstrating robustness with estimation errors.

\begin{figure}[tp]
\center
\includegraphics[width=0.49\textwidth]{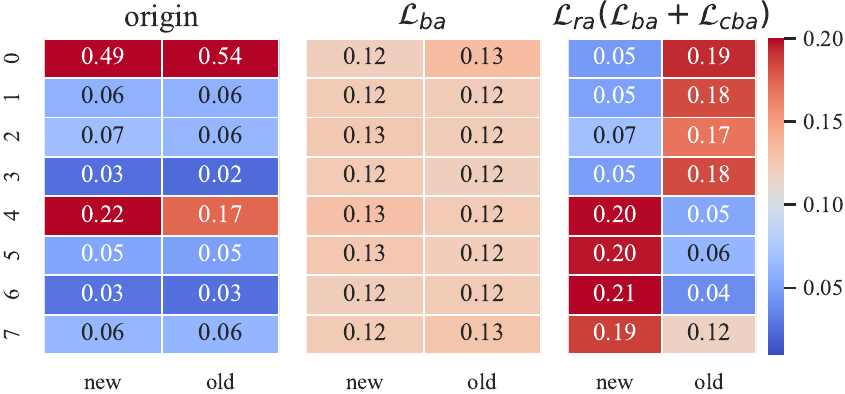}
\caption{{Visualization of the route assignment probability for samples predicted as old and new classes in 8 experts}.} 
\label{fig::route_assign} 
\end{figure}

\emph{6) Visualization of Assignment Probability:}
{To verify the role of the route assignment constraint, we also visualize the assignment probabilities for samples predicted as old and new classes under the original SimGCD loss, the loss with $\mathcal{L}_{ba}$, and the loss with $\mathcal{L}_{ba}+\mathcal{L}_{cba}$ on Aircraft dataset in Fig. \ref{fig::route_assign}. Without any additional loss, the ``winner-takes-all'' phenomenon occurs and the imbalance is greatly alleviated via $\mathcal{L}_{ba}$. Note that a significant portion of samples predicted as old or new classes are distributed to their corresponding experts with the addition of $\mathcal{L}_{ba}$ and $\mathcal{L}_{cba}$. It validates the effectiveness of $\mathcal{L}_{cba}$ to separate the data from old and new classes on the route level. 
}

\begin{figure}[tp]
\center
\includegraphics[width=0.49\textwidth]{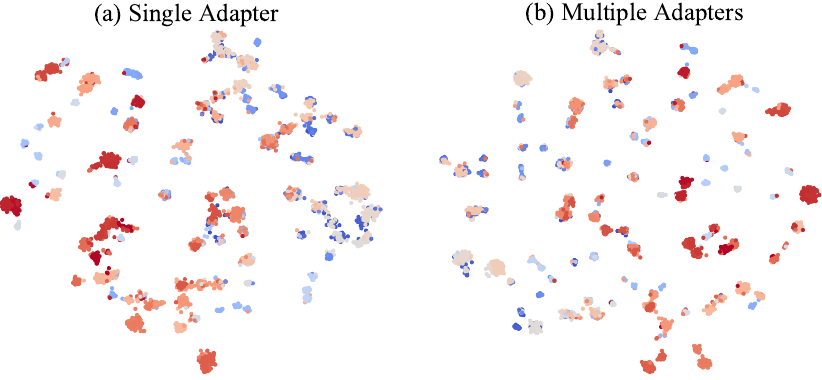}
\caption{$t$-SNE visualization of the proposed methods based on (a) single adapter and (b) multiple adapters. The samples in the old classes are marked in blue while those in new classes are marked in red.} 
\label{fig::tsne} 
\end{figure}

\emph{7) Visualization of $t$-SNE Results of the Features:}  \label{app:tsne}
{To further validate the impact of AdaptGCD on the representations, $t$-SNE \cite{van2008visualizing} is utilized to visualize the features generated by using a single adapter versus multi-expert adapters (with constraints). As shown in Fig. \ref{fig::tsne}, compared to the single adapter, the features generated by the multiple adapters for the new classes (in red) exhibit less interference with the old classes (in blue) and demonstrate greater compactness. These observations confirm the effectiveness of the multi-expert adapters.}

\section{Conclusion} \label{section:conclusion}

In this study, we propose a novel AdaptGCD in generalized category discovery, which is the pioneering work to merge adapter tuning into the GCD framework. It integrates learnable adapters with the fixed backbone to adapt to the downstream GCD task as well as preserves the pretrained priority. Additionally, considering the imbalance between the supervision of the old and new classes, we present a multi-expert adapter structure and a route assignment constraint to separate the data from old and new classes into different expert groups, further reducing mutual interference. The efficacy of the proposed strategy is demonstrated in 7 datasets, including 3 generic datasets 3 fine-grained datasets, and a long-tailed dataset. It can also serve as a plug-and-play module to further improve the performance of advanced GCD methods. Looking forward to future work, a promising direction is to explore more effective fine-tuning strategies in GCD tasks, including the RepAdapter \cite{DBLP:journals/corr/abs-2302-08106}, or the combination of different strategies like GLoRA \cite{DBLP:journals/corr/gloraabs-2306-07967}. In terms of limitations, the proposed method introduces additional parameters, which consume extra computational cost during training and inference. 

\section{Acknowledgement}
The authors are thankful for the financial support by the National Natural Science Foundation of China (U22B2048, 62476274, 62206293).

\bibliographystyle{IEEEtran}
\bibliography{egbib}

\begin{IEEEbiography}
[{\includegraphics[width=0.9in,height=1.1in,clip,keepaspectratio]{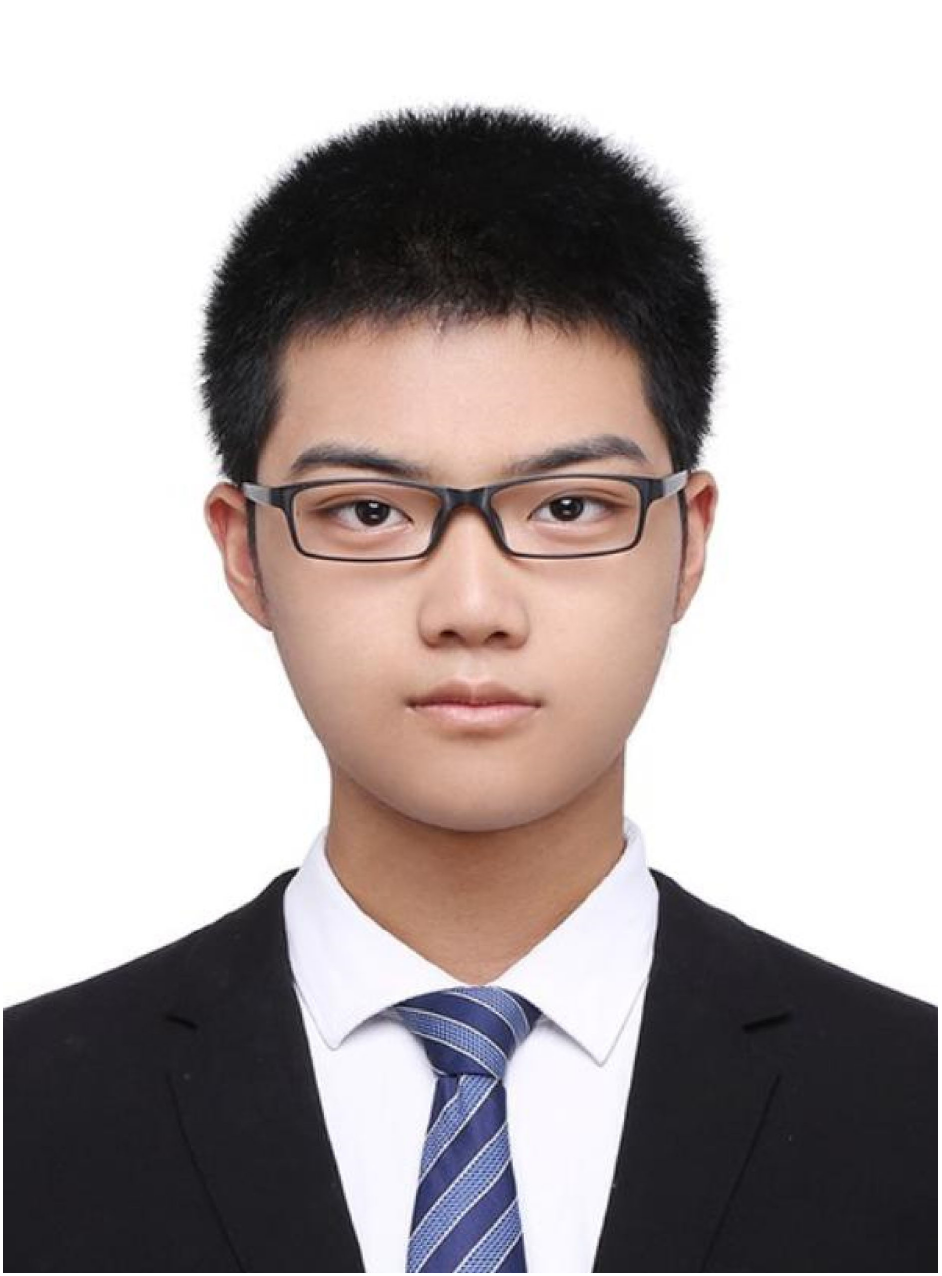}}]{Yuxun Qu} received a B.E. degree from Tianjin University, Tianjin, China, in 2019, and a master degree from Institute of Automation, Chinese Academy of Sciences (CAS), Beijing, China, in 2022. He is currently an Ph.D candidate in the Department of Computer Science, Nankai University. His research interests include artificial intelligence, machine learning, semi-supervised learning, open-world learning, and computer vision.
\end{IEEEbiography}
\begin{IEEEbiography}
[{\includegraphics[width=0.9in,height=1.1in,clip,keepaspectratio]{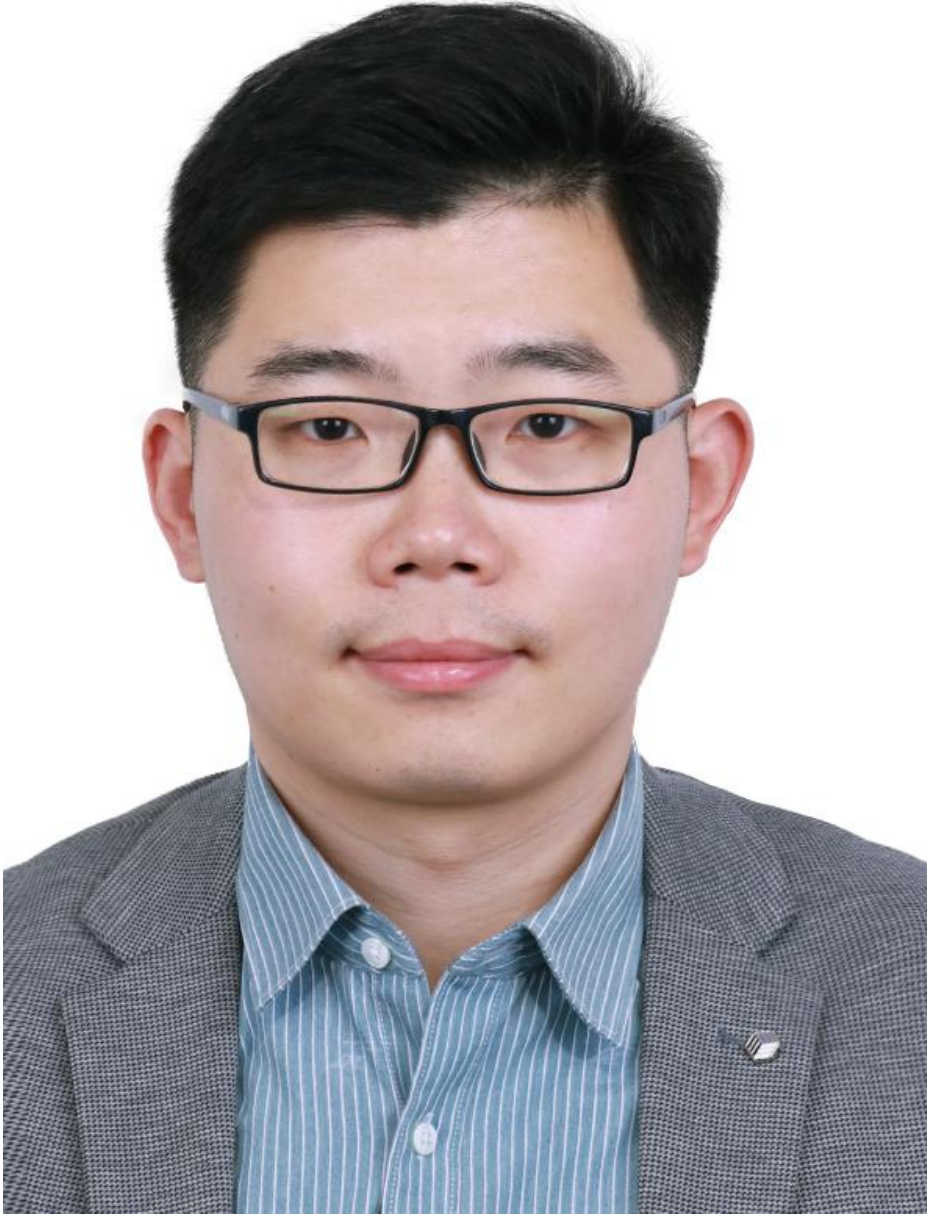}}]{Yongqiang Tang} received the B.S. degree in Department of Automation, Central South University, Changsha, Hunan, China, in 2014 and the Ph.D. degree at the Institute of Automation, Chinese Academy of Sciences (CAS), Beijing, China, in 2019. He is  currently an Associate Professor with the State Key Laboratory of Multimodal Artificial Intelligence Systems, Institute of Automation, CAS. He has published more than 60  papers in major international journals and conferences including the ICML, NeurIPS, SIGKDD, IEEE TPAMI, TIP, TMM, TCYB, TNNLS, TKDE, TGRS, TCSVT, TII, TIM, etc.	His current research interests  include machine learning, computer vision and data mining.
\end{IEEEbiography}
\begin{IEEEbiography}[{\includegraphics[width=0.9in,height=1.1in,clip,keepaspectratio]{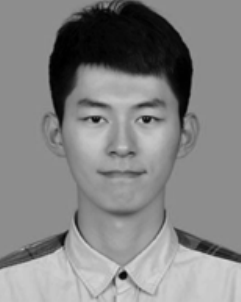}}] {Chenyang Zhang} received the B.E. degree from the Department of Information Science and Engineering, Lanzhou University, Lanzhou, Gansu, China, in 2016, and the Ph.D. degree from the Institute of Automation, Chinese Academy of Sciences (CAS), Beijing, China, in 2021. He is currently an Associate Professor with State Key Laboratory of Multimodal Artificial Intelligence Systems, Institute of Automation, CAS. His research interests include artificial intelligence, machine learning, and computer vision.
\end{IEEEbiography}
\begin{IEEEbiography}[{\includegraphics[width=0.9in,height=1.1in,clip,keepaspectratio]{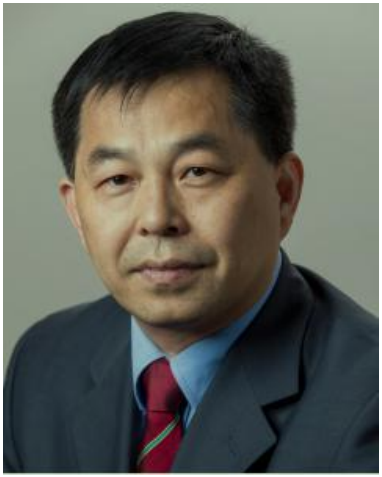}}]{Wensheng Zhang} received his Ph.D. degree in	Pattern Recognition and Intelligent Systems	from the Institute of Automation, Chinese Academy of Sciences (CAS), in 2000. He joined	the Institute of Software, CAS, in 2001. He is a	Professor of Machine Learning and Data Mining	and the Director of Research and Development Department, Institute of Automation, Chinese Academy of Sciences (CAS). He is also with the Guangzhou University. His research interests	include computer vision, pattern recognition and artificial intelligence.
\end{IEEEbiography}

\vfill

\end{document}